%% file: main.tex
\documentclass[sigconf,9pt]{acmart}

\usepackage{microtype}
\usepackage{graphicx}

\usepackage{subcaption}
\usepackage{tabu}
\usepackage{booktabs} 
\usepackage{amsmath}
\usepackage{xspace}
\usepackage{enumitem}
\usepackage{color}
\usepackage{xcolor}
\usepackage{soul}
\usepackage[ruled,linesnumbered]{algorithm2e}
\usepackage{extarrows}
\usepackage{stmaryrd}
\usepackage{gensymb}
\usepackage{lipsum}
\usepackage{tikz}
\usepackage{xcolor}
\usepackage{float}
\usepackage{stfloats} 

\newcommand*\circled[1]{\tikz[baseline=(char.base)]{
            \node[shape=circle,fill,inner sep=1.2pt] (char) {\textcolor{white}{#1}};}}

\newlength\mylenin
\newcommand\myinput[1]{%
\settowidth\mylenin{\KwIn{}}%
\setlength\hangindent{\mylenin}%
\hspace*{\mylenin}#1\\}

\usepackage{textgreek} 

\let\oldnl\nl 
\newcommand{\nonl}{\renewcommand{\nl}{\let\nl\oldnl}} 

\newlength\mylenout

\usepackage{amsmath}
\DeclareMathOperator*{\argmax}{argmax}
\usepackage{amssymb}
\usepackage{dsfont}
\usepackage[bb=boondox]{mathalfa}

\makeatletter
\def\footnoterule{\relax%
  \kern-5pt
  \hbox to \columnwidth{\hfill\vrule width 1\columnwidth height 0.4pt\hfill}
  \kern4.6pt}
\makeatother

\newcommand{\tool}{\texttt{{SPINN}}}
\definecolor{ilias_color_TM}{RGB}{191, 232, 255}
\definecolor{stelios_colour}{RGB}{144, 238, 144}
\definecolor{light_red}{RGB}{255, 204, 204}

\newif\ifcomment

\commentfalse

\ifcomment
\newcommand{\stelios}[1]{\sethlcolor{stelios_colour}\hl{[\textbf{Stelios:} #1]}}
\newcommand{\steve}[1]{\sethlcolor{cyan}\hl{[\textbf{Steve:} #1]}}
\newcommand{\manote}[1]{\sethlcolor{pink}\hl{[\textbf{Mario:} #1]}}
\newcommand{\il}[1]{\sethlcolor{ilias_color_TM}\hl{[\textbf{Ilias:} #1]}}
\newcommand{\cut}[1]{\sethlcolor{light_red}\hl{[#1]}}
\newcommand{\blue}[1]{\textcolor{blue}{#1}} %
\else
\newcommand{\stelios}[1]{}
\newcommand{\steve}[1]{}
\newcommand{\il}[1]{}
\newcommand{\manote}[1]{}
\newcommand{\cut}[1]{}
\newcommand{\blue}[1]{#1}
\fi

\newif\ifacmversion
\acmversionfalse

\ifacmversion
\copyrightyear{2020} 
\acmYear{2020} 
\setcopyright{acmcopyright}\acmConference[MobiCom '20]{The 26th Annual International Conference on Mobile Computing and Networking}{September 21--25, 2020}{London, United Kingdom}
\acmBooktitle{The 26th Annual International Conference on Mobile Computing and Networking (MobiCom '20), September 21--25, 2020, London, United Kingdom}
\acmPrice{15.00}
\acmDOI{10.1145/3372224.3419194}
\acmISBN{978-1-4503-7085-1/20/09}
\usepackage{natbib}

\else

\usepackage{background}
\backgroundsetup{angle=0,
    scale=1,
    color=black,
    firstpage=true,
    position=current page.north,
    hshift=0pt,
    vshift=-20pt,
    contents={\ifnum\value{page}=1 PREPRINT: Accepted at the 26th Annual International Conference on Mobile Computing and Networking (MobiCom), 2020 \else \fi}
}
\fi


\ifacmversion
\else
\renewcommand\footnotetextcopyrightpermission[1]{} 
\setcopyright{none}
\fi

\title{SPINN: Synergistic Progressive Inference \\of Neural Networks over Device and Cloud}

\author{
{Stefanos Laskaridis$^\dagger$*, Stylianos I. Venieris$^\dagger$*, \mbox{Mario Almeida$^\dagger$*, Ilias Leontiadis$^\dagger$*, Nicholas D. Lane}$^{\dagger,\ddagger}$}}

\affiliation{\institution{$^\dagger$Samsung AI Center, Cambridge\hspace{+0.75cm}$^\ddagger$University of Cambridge}{\Small\textit{{* Indicates equal contribution.}}}}

\email{{stefanos.l, s.venieris, mario.a, i.leontiadis, nic.lane}@samsung.com}

\authorwithoutinstuition{{Stefanos Laskaridis, Stylianos I. Venieris, Mario Almeida, Ilias Leontiadis, Nicholas D. Lane}}

\input{sections/abstract.tex}

\begin{document}

\fancyhead{}
\maketitle

\input{sections/introduction.tex}
\input{sections/background.tex}
\input{sections/architecture.tex}

\input{sections/evaluation.tex}
\input{sections/limitations.tex}

\input{sections/conclusion.tex}

\balance
\bibliographystyle{ACM-Reference-Format}
\bibliography{references}

\end{document}

%% file: sections/abstract.tex
\begin{abstract}
Despite the soaring use of convolutional neural networks (CNNs) in mobile applications, uniformly sustaining high-performance inference on mobile has been elusive due to the excessive computational demands of modern CNNs and the increasing diversity of deployed devices. 
A popular alternative comprises offloading CNN processing to powerful cloud-based servers. 
Nevertheless, 
by relying on the cloud to produce outputs, emerging mission-critical and high-mobility applications, such as drone obstacle avoidance or interactive applications, can suffer from the dynamic connectivity conditions and the uncertain availability of the cloud.
In this paper, we propose \tool{}, a distributed inference system that employs synergistic device-cloud computation 
together with a progressive inference method to deliver fast and robust CNN inference across diverse settings.
The proposed system introduces a novel scheduler that co-optimises the early-exit policy and the CNN splitting at run time, in order to adapt to dynamic conditions and meet user-defined service-level requirements. 
Quantitative evaluation illustrates that \tool{} outperforms its state-of-the-art collaborative inference counterparts by up to 2$\times$ in achieved throughput under varying network conditions, reduces the server cost by up to 6.8$\times$ and improves accuracy by 20.7\% under latency constraints, 
while providing robust operation under uncertain connectivity conditions and significant energy savings compared to cloud-centric execution.
\end{abstract}

%% file: sections/introduction.tex


\section{Introduction}
\label{sec:intro}



With the spectrum of CNN-driven applications expanding rapidly, their deployment across mobile platforms poses significant challenges.
%
%
Modern CNNs~\cite{He_2016,Szegedy_2017} have excessive computational demands that hinder their wide adoption in resource-constrained mobile devices.
Furthermore, emerging user-facing~\cite{facebook2019} and mission-critical~\cite{trailnet_2017,learn_to_fly_2018,adas_2018} CNN applications require low-latency processing to ensure high quality of experience (QoE)~\cite{Cartas_2019} and safety~\cite{adas_mit_2019}.

\begin{figure}[t]
      \centering
      \includegraphics[trim={4.15cm 3.25cm 5.15cm 3cm},clip,width=.975\columnwidth]{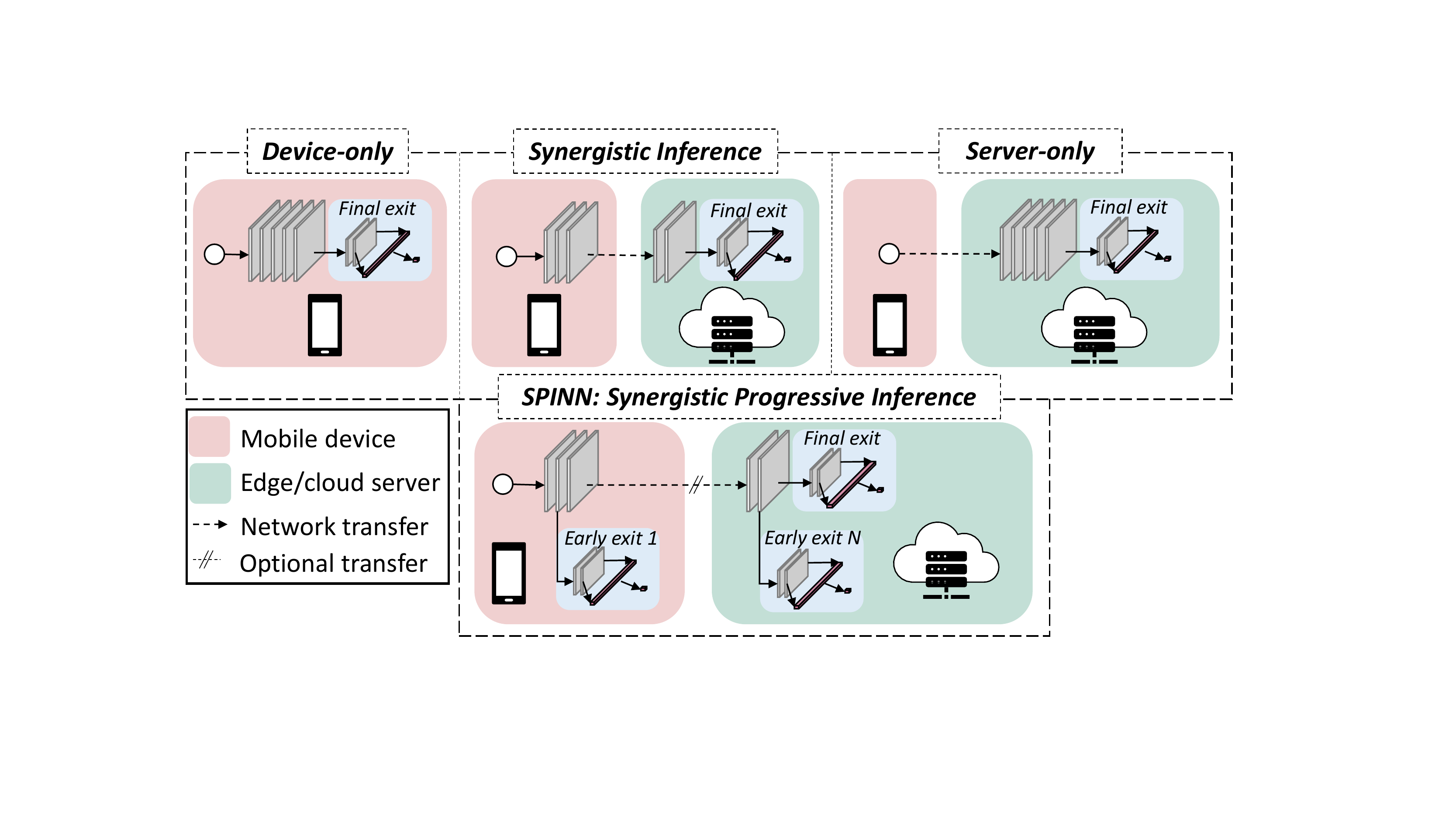}
      \vspace{-0.4cm}
      \caption{Existing methods vs. \tool{}.}
      \vspace{-0.4cm}
      \label{fig:proposed_system}
\end{figure}

Given the recent trend of integrating powerful System-on-Chips (SoCs) in consumer devices~\cite{ai_benchmark_2019,embench_2019,wang_2020}, direct on-device CNN execution is becoming possible (Figure~\ref{fig:proposed_system} - top left).
Nevertheless, while flagship devices can support the performance requirements of CNN workloads, the current landscape is still very diverse, including previous-gen and low-end models~\cite{facebook2019}.
In this context, the less powerful low-tier devices struggle to consistently meet the application-level performance needs~\cite{embench_2019}.

As an alternative,
service providers typically employ cloud-centric solutions (Figure~\ref{fig:proposed_system} - top right). With this setup, inputs collected by mobile devices are transmitted to a remote server to perform CNN inference 
using powerful accelerators~\cite{facebook_datacenter_2018,dl_xeon_2018,Jouppi2017,amazon_inferentia,baidu_dnns_2014,microsoft2018}. 
However, this extra computation capability comes at a price. First, cloud execution is highly dependent on the dynamic network conditions, with performance dropping radically when the communication channel is degraded.
Second, hosting resources capable of accelerating machine learning tasks comes at a significant cost 
~\cite{kozyrakis_2013}. 
Moreover, while  public cloud providers offer elastic cost scaling, there are also privacy and security concerns~\cite{singh2017cloud}. 




To address these limitations, a recent line of work~\cite{Kang2017,Hu2019,edgent_2020} has proposed the collaboration between device and cloud for CNN inference (Figure~\ref{fig:proposed_system} - top center). Such schemes typically treat the CNN as a computation graph and partition it between device and cloud. At run time, the client executes the first part of the model and transmits the intermediate results to a remote server. The server continues the model execution and returns the final result back to the device. Overall, this approach allows tuning the fraction of the CNN that will be executed on each platform based on \mbox{their capabilities}.

Despite their advantages, existing device-cloud collaborative inference solutions suffer from a set of limitations. 
First, similar to cloud execution, the QoE is greatly affected by the  network conditions as execution can fail catastrophically when the link is severely deteriorated.
This lack of network fault tolerance also prevents the use of more cost-efficient cloud solutions, \textit{e.g.} using ephemeral spare cloud resources at a fraction of the price.\footnote{AWS Spot Instances -- \url{https://aws.amazon.com/ec2/spot/}.} 
Furthermore, CNNs are increasingly deployed in applications with stringent demands across multiple dimensions (\textit{e.g.} target latency, throughput and accuracy, or device and cloud costs).\footnote{Typically expressed as service-level agreements (SLAs).} Existing collaborative methods cannot sufficiently meet these requirements.
To this end, we present \tool{}, a distributed system that enables robust CNN inference in highly dynamic environments, while meeting multi-objective application-level requirements (SLAs).
This is accomplished through a novel scheduler that takes advantage of progressive inference; a mechanism that allows the system to exit early at different parts of the CNN during inference, based on the input complexity (Figure~\ref{fig:proposed_system}~-~bottom).
The scheduler optimises the overall execution by jointly tuning both the split point selection and the early-exit policy at run time to sustain high performance and meet the application SLAs under fluctuating resources (\textit{e.g.}~network speed, device/server load).
The guarantee of a local early exit renders server availability non-critical and enables robust operation even under uncertain connectivity.
Overall, this work makes the following key contributions:
\vspace{-0.1cm}
\begin{itemize}[leftmargin=*]
    \item \blue{A progressive inference mechanism that enables the fast and reliable execution of CNN inference across device and cloud.
    Concretely, on top of existing early-exit designs, we propose an early-exit-aware cancellation mechanism that allows the interruption of the (local/remote) inference when having a confident early prediction, thus minimising redundant computation and transfers during inference. 
    Simultaneously, reflecting on the uncertain connectivity of mobile devices we design an early-exit scheme with robust execution in mind, even under severe connectivity disruption or cloud unavailability. By carefully placing the early exits in the backbone network and allowing for graceful fallback to locally available results, we guarantee the responsiveness and reliability of the system and overcome limitations of existing offloading systems.}
    
    \item A CNN-specific packing mechanism that exploits the reduced-precision resilience and sparsity of CNN workloads to minimise transfer overhead. \blue{Our communication optimiser combines a lossless and an accuracy-aware lossy compression component which exposes previously unattainable designs for collaborative inference, while not sacrificing the end accuracy of the system.}
    
    \item An SLA- and condition-aware scheduler that co-optimises i) the early-exit policy of progressive CNNs and ii) their partitioning between device and cloud at run time. The proposed scheduler employs a multi-objective framework to capture the user-defined importance of multiple performance metrics and translate them into SLAs. Moreover, by surveilling the volatile network conditions and resources load at run time, the scheduler dynamically selects the configuration that yields the highest performance \blue{by taking into account contextual runtime information and feedback from previous executions}.
\end{itemize}

%% file: sections/background.tex
\vspace{-0.25cm}
\section{Background and Related Work}
\label{sec:background}


To optimise the execution of CNN workloads, several solutions have been proposed, from compiler~\cite{tensorflow_2016,nn_compiler_2018,astra_tool_2019} and runtime optimisations~\cite{deepx_2016,ulayer_2019,liu_2019} to custom cloud~\cite{Kang2017,brainwave_2018,facebook_datacenter_2018} and accelerator designs~\cite{sys_array_2017,fpgaconvnet_2018}. While these works target a single model with device- or cloud-only execution, the increased computational capabilities of client devices~\cite{ai_benchmark_2019,embench_2019} have led to schemes that maximise performance via device-cloud synergy.
Next, we outline significant work in this direction and visit approximate computing alternatives which exploit accuracy-latency trade-offs during inference.



\textbf{Approximate Inference.}
%
%
%
In applications that can tolerate some accuracy drop, a line of work~\cite{Han:2016:MAE:2906388.2906396, nestdnn_2018, Lee_2019} exploits the accuracy-latency trade-off through various techniques. In particular, \texttt{NestDNN} \cite{nestdnn_2018} employs a multi-capacity model that incorporates multiple descendant (\textit{i.e.} pruned) models to expose an accuracy-complexity trade-off mechanism. However, such models cannot be natively split between device and cloud.
%
On the other hand, model selection systems \cite{Han:2016:MAE:2906388.2906396,Lee_2019} 
employ multiple variants of a single model (\textit{e.g.}~quantised, pruned) with different accuracy-latency trade-offs. At run time, they choose the most appropriate variant based on the application requirements and determine where it will be executed.
Similarly, classifier cascades~\cite{noscope_2017,focus_2018,taylor_2018,cascadecnn2018,cascadecnn2020date} require multiple models to obtain performance gains. Despite the advantages of both,
using multiple models adds substantial overhead in terms of maintenance, \mbox{training and deployment}.

 
\textbf{Progressive Inference Networks.} 
\blue{A growing body of work from both the research~\cite{branchynet2016,Huang2017,sdn_icml_2019,scan2019neurips,deebert2020acl} and industry communities~\cite{testorrent,nervana} has proposed transforming a given model into a progressive inference network by introducing intermediate exits throughout its depth.}
By exploiting the different complexity of incoming samples, easier examples can early-exit and save on further computations.
So far, existing works have mainly explored the hand-crafted design of early-exit architectures (\texttt{MSDNet}~\cite{Huang2017}, \blue{\texttt{SCAN}~\cite{scan2019neurips}}), the platform- and SLA-agnostic derivation of early-exit networks from generic models (\texttt{BranchyNet}~\cite{branchynet2016}, \texttt{SDN}~\cite{sdn_icml_2019}) \blue{or the hardware-aware deployment of such networks (\texttt{HAPI}~\cite{hapi2020iccad})}.
Despite the recent progress, these techniques have not capitalised upon the unique potential of such models to yield high mobile performance through distributed execution and app-tailored early-exiting. In this context, \tool{} is the first progressive inference approach equipped with a principled method of selectively splitting execution between device and server, while also tuning the early-exit policy, enabling high performance across dynamic settings.



\textbf{Device-Cloud Synergy for CNN Inference.}
To achieve efficient CNN processing, several works have explored collaborative computation over device, edge and cloud. 
One of the most prominent pieces of work, \texttt{Neurosurgeon}~\cite{Kang2017}, partitions the CNN between a device-mapped \textit{head} and a cloud-mapped \textit{tail} and selects a single split point based on the device and cloud load as well as the network conditions. Similarly, \texttt{DADS}~\cite{Hu2019} tackles CNN offloading, but from a scheduler-centric standpoint, with the aim to yield the optimal partitioning scheme in the case of high and low server load. However, both systems only optimise for single-criterion objectives (latency or energy consumption), they lack support for app-specific SLAs, and suffer catastrophically when the remote server is unavailable.
With a focus on data transfer, \texttt{JALAD}~\cite{Li2019} incorporates lossy compression to minimise the offload transmission overhead. Nevertheless, to yield high performance, the proposed system sacrifices substantial accuracy (\textit{i.e.}~>5\%).
\texttt{JointDNN}~\cite{jointdnn_2019} modelled CNN offloading as a graph split problem, but targets only offline scheduling and static environments instead of highly dynamic mobile settings. 
Contrary to these systems, \tool{} introduces a novel scheduler that adapts the execution to the dynamic contextual conditions and jointly tunes the offloading point and early-exit policy to meet the application-level requirements. Moreover, by guaranteeing the presence of a local result, \tool{} provides resilience to server \mbox{disconnections}.

Apart from offloading CNNs to a dedicated server, a number of works have focused on tangential problems. 
\texttt{IONN}~\cite{Jeong2018} tackles a slightly different problem, where instead of preinstalling the CNN model to a remote machine, the client device can offload to any close-by server by transmitting both the incoming data and the model in a shared-nothing setup.
Simultaneously, various works~\cite{Mao2017,Mao2017a,Zhao2018} have examined the case where the client device can offload to other devices in the local network.
Last, \cite{7979979} also employs cloud-device synergy and progressive inference, but with a very different focus, \textit{i.e.} to perform joint classification from a multi-view, multi-camera standpoint. Its models, though, are statically allocated to devices and its fixed, statically-defined early-exit policy, renders it impractical for dynamic environments.

\begin{figure*}[t]
    \centering
    \includegraphics[trim={0cm 6cm 0cm 6cm},clip,width=0.8\textwidth]{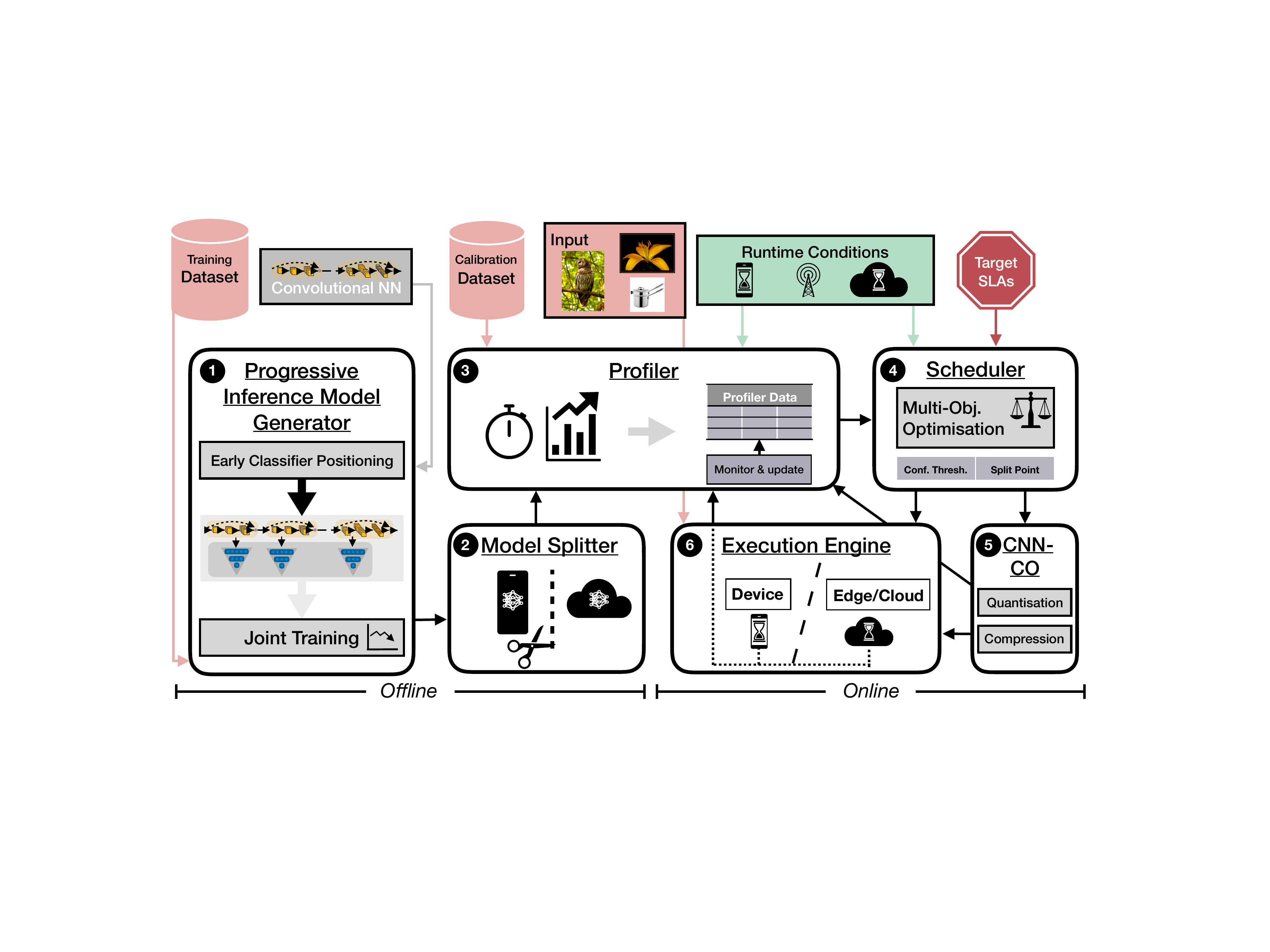}
    \vspace{-0.8cm}
    \caption{Overview of \tool{}'s architecture.}
    \vspace{-0.4cm}
    \label{fig:dope-nn-arch}
\end{figure*}

\textbf{Offloading Multi-exit Models.}
Closer to our approach, \texttt{Edgent} \cite{edgent_2020} proposes a way of merging offloading with multi-exit models. 
Nonetheless, this work has several limitations.
First, the inference workflow disregards data locality and always starts from the cloud. Consequently, inputs are always transmitted, paying an additional transfer cost.
Second, early-exit networks are not utilised with progressive inference, \textit{i.e.} inputs do not early-exit based on their complexity. 
Instead, \texttt{Edgent} tunes the model's complexity by selecting a \textit{single} intermediary exit for all inputs. Therefore, the end system does not benefit from the variable complexity of inputs.
Finally, the system has been evaluated solely on simple models (AlexNet) and datasets (CIFAR-10), less impacted by low-latency or unreliable network conditions.
In contrast, \tool{} exploits the fact that data already reside on the device to avoid wasteful input transfers, and employs a CNN-tailored technique to compress the offloaded data. Furthermore, not only our scheduler supports additional optimisation objectives, but it also takes advantage of the input's complexity to exit early, saving resource usage with minimal impact on accuracy.

%% file: sections/architecture.tex
\vspace{-0.5em}
\section{Proposed System}
\label{sec:methodology}


To remedy the limitations of existing systems, \tool{} employs a progressive inference approach to alleviate the hard requirement for reliable device-server communication. 
The proposed system introduces a scheme of distributing progressive early-exit models across device and server, in which one exit is always present on-device, guaranteeing the availability of a result at all times.
Moreover, as early exits along the CNN provide varying levels of accuracy, \tool{} casts the acceptable prediction confidence as a tunable parameter to adapt its accuracy-speed trade-off.  
Alongside, we propose a novel run-time scheduler that jointly tunes the split point and early-exit policy of the progressive model, yielding a deployment tailored to the application performance requirements under dynamic conditions.
Next, we present \tool{}'s high-level flow followed by a description of its components.


\vspace{-0.5em}
\subsection{Overview}
\label{sec:overview}

\tool{} comprises offline components, run once before deployment, and online components, which operate at run time. Figure \ref{fig:dope-nn-arch} shows a high-level view of \tool{}. 
Before deployment, \tool{} obtains a CNN model and derives a progressive inference network. This is accomplished by introducing early exits along its architecture and jointly training them using the supplied training set (Section \ref{sec:prog_inf} \circled{1}). Next, the model splitter component (Section~\ref{sec:partitioning}  \circled{2}) identifies all candidate points in the model where computation can be split between device and cloud.
Subsequently, the offline profiler (Section~\ref{sec:profiler} \circled{3}) calculates the exit-rate behaviour of the generated progressive model as well as the accuracy of each classifier. Moreover, it measures its performance on the client and server, serving as initial inference latency estimates.

At run time, the scheduler (Section \ref{sec:opt_framework} \circled{4}) obtains these initial timings along with the target SLAs and run-time conditions and decides on the optimal split and early-exit policy. Given a split point, the communication optimiser (Section~\ref{sec:comms_opt} \circled{5}) exploits the CNN's sparsity and resilience to reduced bitwidth to compress the data transfer and increase the bandwidth utilisation. The execution engine (Section~\ref{sec:offload} \circled{6}) then orchestrates the distributed inference execution, handling all communication between partitions. Simultaneously, the online profiler monitors the execution across inferences, as well as the contextual factors (\textit{e.g.} network, device/server load) and updates the initial latency estimates. This way, the system can adapt to the rapidly-changing environment, reconfigure its execution and maintain the same QoE.

\vspace{-0.2cm}
\subsection{Progressive Inference Model Generator}
\label{sec:prog_inf}

Given a CNN model, \tool{} derives a progressive inference network (Figure~\ref{fig:dope-nn-arch}~\circled{1}). This process comprises a number of key design decisions:
{1)}~the \textit{number}, \textit{position} and  \textit{architecture}  of intermediate classifiers (early exits), {2)}~the \textit{training scheme} and {3)}~the \mbox{\textit{early-exit policy}}. 

\textbf{Early Exits.}
We place the intermediate classifiers along the depth of the architecture with equal distance in terms of FLOP count. 
With this platform-agnostic positioning strategy, we are able to obtain a progressive inference model that supports a wide range of latency budgets while being portable across devices. 
With respect to their number, we introduce six early exits in order to guarantee their convergence when trained jointly~\cite{sdn_icml_2019,li2019improved}, placed at 15\%, 30\%, \dots 90\% of the network's total FLOPs.
Last, we treat the architecture of the early exits as an invariant, adopting the design of~\cite{Huang2017}, so that all exits have the same expressivity~\cite{exp_icml_2017}.

\begin{figure}[t]
  \centering
  \vspace{-0.3cm}
  \includegraphics[trim={2.2cm 0cm 2.2cm 0cm},clip,width=0.99\columnwidth]{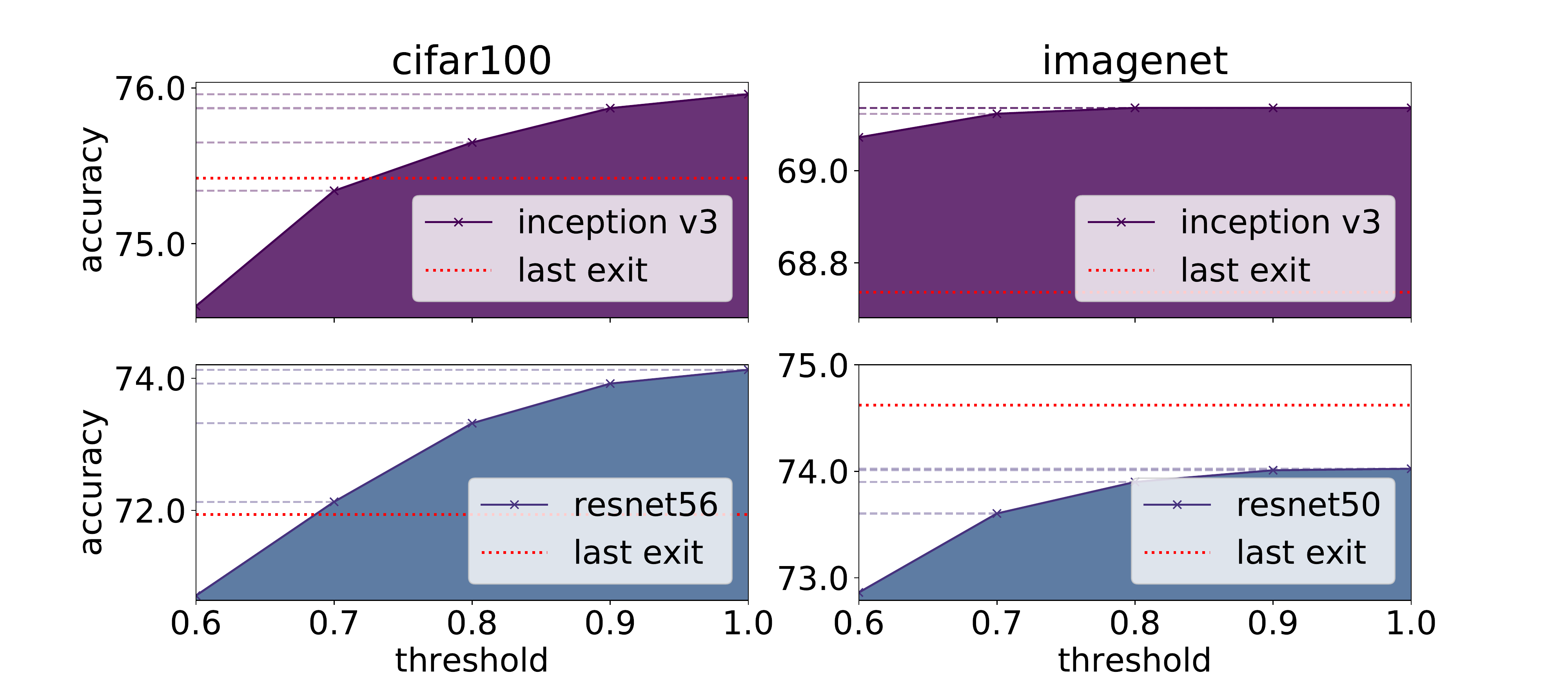}
  \vspace{-0.45cm}
  \caption{Accuracy of progressive networks across \mbox{different} confidence threshold values.}
  \vspace{-0.45cm}
  \label{fig:accuracy_conf_theshold}
\end{figure}

\textbf{Training Scheme.}
We jointly train all classifiers from scratch and employ the cost function introduced in~\cite{sdn_icml_2019} as follows: $\mathcal{L} = \sum_{i=0}^{N-1}{\tau_i*\mathcal{L}_i}$ 
%
%
with $\tau_i$ starting uniformly at $0.01$ and linearly increasing it to a maximum value of $C_i$, which is the relative position of the classifier in the network ($C_0 = 0.15$, $C_1 = 0.3$, \dots, $C_{\text{final}}=1$). 
The rationale behind this is to address the problem of ``overthinking''~\cite{sdn_icml_2019}, where some samples can be correctly classified by early exits while being misclassified deeper on in the network. This scheme requires the fixed placement of early exits prior to the training stage. 
Despite the inflexibility of this approach to search over different early-exit placements, it yields higher accuracy compared to the two-staged approach of training the main network and the early classifiers in isolation. \blue{In terms of training time, the end-to-end early-exit networks can take from 1.2$\times$ to 2.5$\times$ the time of the original network training, depending on the architecture and number of exits. 
In fact, the higher training overhead happens when the ratio of $\frac{FLOPS_{early\_classifier}}{FLOPS_{original\_network}}$ is higher. 
However, given that this cost is paid once offline, it is quickly amortised by the runtime latency benefits of early exiting on confident samples.}


\textbf{Early-exit Policy.} 
For the early-exit strategy, we estimate a classifier's \textit{confidence} for a given input using the top-1 output value of its softmax layer (Eq.~(\ref{eq:softmax}))~\cite{Guo2017}.
An input takes the i-th exit if the prediction confidence is higher than a tunable threshold, $thr_{\text{conf}}$ (Eq.~(\ref{eq:conf_threshold})).
The exact value of $thr_{\text{conf}}$ provides a trade-off between the latency and accuracy of the progressive model and determines the \textit{early-exit policy}.
At run time, \tool{}'s scheduler periodically tunes $thr_{\text{conf}}$ to customise the execution to the application's needs. 
If none of the classifiers reaches the confidence threshold, the most confident among them is used as the output prediction (Eq.~(\ref{eq:max_conf})).
\vspace{-0.2cm}
\begin{small}
    \begin{align}
        &\text{softmax}(z)_i = \frac{e^{z_i}}{\sum_{j=1}^{K}e^{z_j}} & ~~~\text{ (\textit{Softmax of i-th exit})} \label{eq:softmax}\\
        &\arg_i\{\max_i\{\text{softmax}_i\} > {thr}_{\text{conf}}\} & ~~~\text{ (\textit{Check i-th exit's top-1})} \label{eq:conf_threshold}\\
        &\argmax_{j\in\text{classifiers}}\{\max_i\{\text{softmax}_i^j\}\} & ~~~\text{ (\textit{Return most confident})}  \label{eq:max_conf}
    \end{align}
\end{small}
where $z_i$ is the output of the final fully-connected layer for the i-th label, $K$ the total number of labels, $j$$\in$$[0,6]$ the classifier index and ${thr}_{\text{conf}}$ the tunable confidence threshold.

\begin{figure}[t]
  \centering
  \includegraphics[width=0.95\columnwidth]{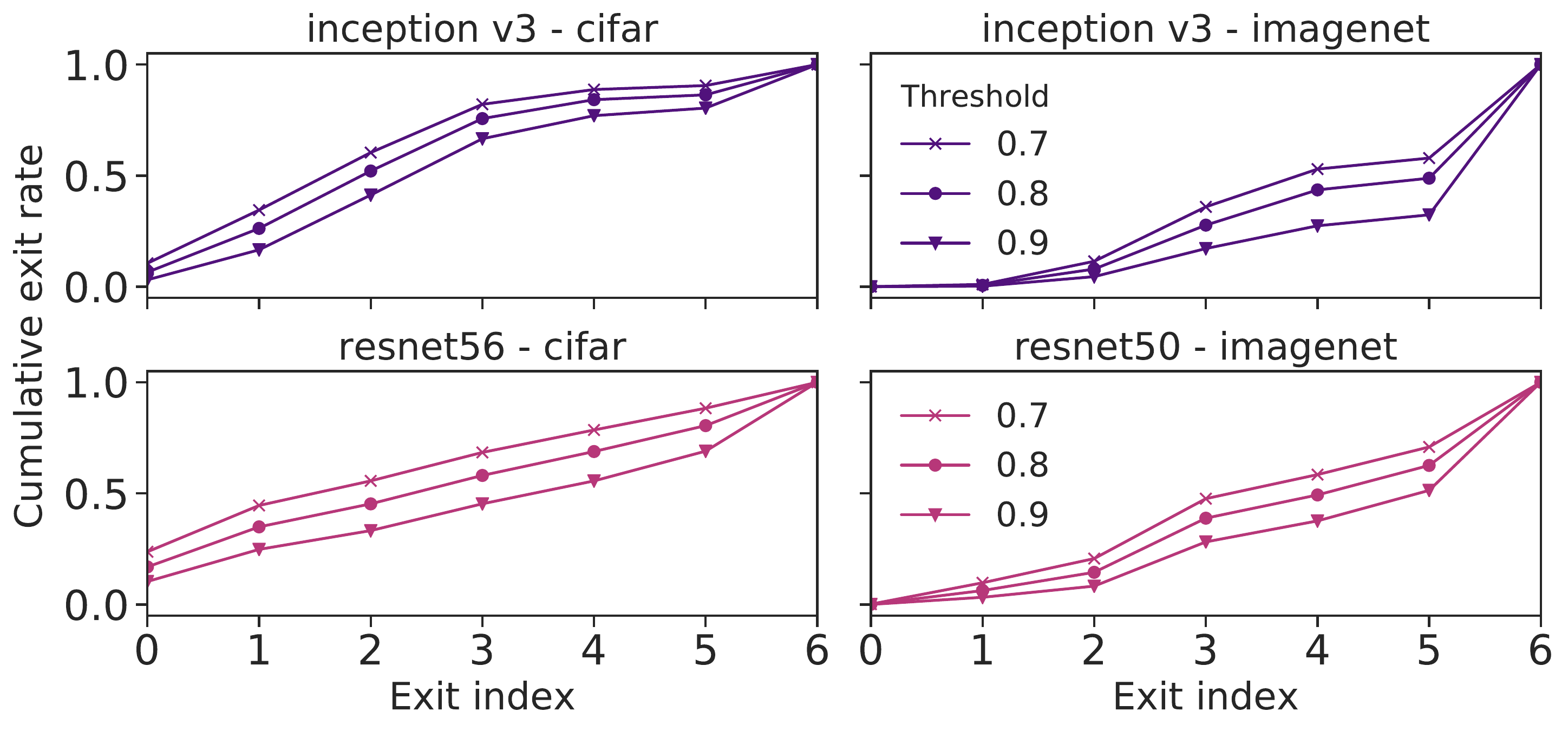}
  \vspace{-0.5cm}
  \caption{Early-exit CDF across different confidence threshold values.}
  \vspace{-0.6cm}
  \label{fig:early_exit_cumsum}
\end{figure}

\textbf{Impact of Confidence Threshold.}
Figure~\ref{fig:accuracy_conf_theshold} and~\ref{fig:early_exit_cumsum} illustrate the impact of different \textit{early-exit policies} on the accuracy and early-exit rate of progressive models, by varying the confidence threshold ($thr_{\text{conf}}$).
%
Additionally, Figure~\ref{fig:accuracy_conf_theshold} reports on the accuracy without progressive inference (\textit{i.e.} last exit only, represented by the red dotted line). 
Note that exiting only at the last exit can lead to lower accuracy than the progressive models for some architectures, a phenomenon that can be attributed to the problem of ``overthinking"\footnote{\blue{``Overthinking"~\cite{sdn_icml_2019} dictates that certain samples that would normally get misclassified by reaching the final classifier of the network if they exit early, they get classified correctly. This leads to small accuracy benefits of progressive inference networks that neither the original model would have (due to early-exiting) nor a single-exit smaller variant (due to late-exiting).
}}. 

Based on the figures, we draw two major conclusions that guide the design of \tool{}'s scheduler.
First, across all networks, we observe a monotonous trend with higher thresholds leading to higher accuracies (Figure~\ref{fig:accuracy_conf_theshold}) while lower ones lead to more samples exiting earlier (Figure~\ref{fig:early_exit_cumsum}). 
This exposes the confidence threshold as a tunable parameter to control accuracy and overall processing time.
Second, different networks behave differently, producing \textit{confident} predictions at different exits along the architecture.
For example, Inception-v3 on CIFAR-100 can obtain a confident prediction earlier on, whereas ResNet-50 on ImageNet cannot classify robustly from early features only. 
In this respect, we conclude that optimising the confidence threshold for each CNN explicitly is key for tailoring the deployment to the target requirements.

\subsection{Model Splitter}
\label{sec:partitioning}

%

After deriving a progressive model from the original CNN, \tool{} aims to split its execution across a client and a server in order to dynamically utilise remote resources as required by the application SLAs.
The model splitter (Figure~\ref{fig:dope-nn-arch}~\circled{2}) is responsible for 1)~defining the potential split points and 2)~identifying them automatically in the given CNN.


\textbf{Split Point Decision Space.}
CNNs typically consist of a sequence of layers, organised in a feed-forward topology. \tool{} adopts a partition scheme which allows splitting the model along its depth at layer granularity. 
%
For a CNN with $N_L$ layers, 
there are $N_L$$-$$1$ candidate points, leading to $2^{N_L-1}$ possible partitions.
To reduce the search space and minimise the number of transmissions across the network, we make two key observations.
First, since CNN final outputs are rather small, once execution is offloaded to the powerful remote server, there is no gain in having two or more split points as this would incur in extra communication costs. 
Second, different layer splits have varying transmission costs and compression potential.
For example, activation layers such as \texttt{ReLU}~\cite{nair2010rectified} cap negative values to zero, which means that their output becomes more compressible~\cite{xu2015empirical,cdma_2018,nikolic2019characterizing} and they can be more efficiently transferred by \tool{}'s \textit{communication optimiser} (Section~\ref{sec:comms_opt}).
Therefore, while \tool{}'s design supports an arbitrary number of split points and layers, in this work, we allow one split point per CNN and reduce the candidate split points to \texttt{ReLU} layers.

\textbf{Automatic Split Point Identification.}
To automatically detect all candidate split points in the given CNN, the model splitter employs a dynamic analysis approach. This is performed by first constructing the execution graph of the model in the target framework (\textit{e.g.} \textit{PyTorch}), identifying all split points and the associated dependencies, and then applying \tool{}'s partitioning scheme to yield the final pruned split point space. 
The resulting set of points defines the allowed partitions that can be selected by the \mbox{scheduler (Section~\ref{sec:opt_framework})}.

\begin{figure}[t]
  \centering
  \includegraphics[width=\columnwidth]{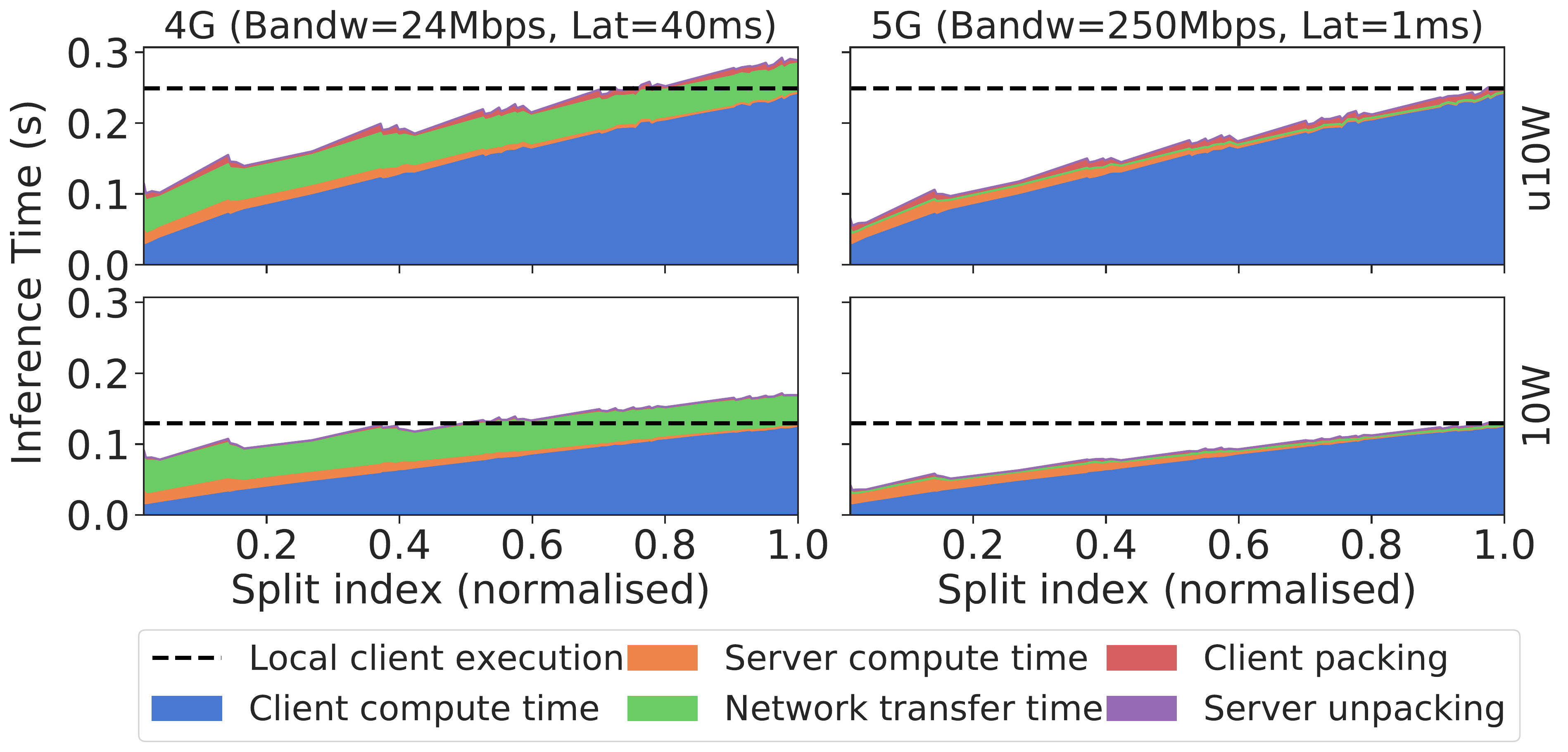}
  \vspace{-0.4cm}
  \caption{ResNet-56 inference times for different \mbox{network} conditions (4G and 5G) and client device \mbox{compute} capabilities (Jetson 10W and u10W), when \mbox{offloading} to the cloud without early exits.}
  \vspace{-0.4cm}
  \label{fig:jetson-times}
\end{figure}


\textbf{Impact of Split Point.}
To investigate how the split point selection affects the overall latency, we run multiple CNN splits between a Nvidia Jetson Xavier client and a server (experimental setup detailed in Section~\ref{sec:eval}).
Figure \ref{fig:jetson-times} shows the breakdown of ResNet-56's inference times with CIFAR-100 over distinct network conditions and client compute capabilities - u10W and 10W. 
Packing refers to the runtime of the communication optimiser module, \mbox{detailed in Section~\ref{sec:comms_opt}}.

Based on the figure, we make two observations. 
First, different split points yield varying trade-offs in client, server and transfer time.
For example, earlier splits execute more on the server, while later ones execute more on-device but often with smaller transfer requirements\footnote{\blue{Note that independently of the amount of transmitted data, there is always a network latency overhead that must be amortised, 
which in the case of 4G, is quite significant.}}. 
This indicates that \tool{}'s scheduler can selectively choose one to minimise any given runtime (\textit{e.g.} device, server or transfer), as required by the user. 
Second, dynamic conditions such as the connectivity, and the device compute capabilities play an important role in shaping the split point latency characteristics illustrated in Figure~\ref{fig:jetson-times}. For example, a lower-end client (u10W) or a loaded server would require longer to execute its allocated split, while low bandwidth can increase the transfer time.
This indicates that it is hard to statically identify the best split point and highlights the need for an informed partitioning that adapts to the environmental conditions in order to meet the application-level performance requirements.

\subsection{Profiler}
\label{sec:profiler}

Given the varying trade-offs of different split points and confidence thresholds, \tool{} considers the client and server load, the networking conditions and the expected accuracy in order to select the most suitable configuration. 
%
%
To estimate this set of parameters, the \textit{profiler} (Figure~\ref{fig:dope-nn-arch}~\circled{3}) operates in two stages: i) offline \mbox{and ii) run-time}.


\textbf{Offline stage:} In the offline phase, the profiler performs two kinds of measurements, \textit{device-independent} and \textit{device-specific}. The former include CNN-specific metrics, such as 1) the size of data to be transmitted for each candidate split and 2) the average accuracy of the progressive CNN for different confidence thresholds. These are measured only once prior to deployment.
Next, the profiler needs to obtain latencies estimates that are specific to each device. To this end, the profiler measures the average execution time per layer by passing the CNN through a calibration set -- sampled from the validation set of the target task. The results serve as the initial latency and throughput estimates.

\textbf{Run-time stage:} At run time, the profiler refines its offline estimates by regularly monitoring the device and server load, as well as the connectivity conditions. 
To keep the profiler lightweight, instead of adopting a more accurate but prohibitively expensive estimator, we employ a 2-staged linear model to estimate the inference latency on-device. 

In the first step, the profiler measures the actual on-device execution time up to the split point $s$, denoted by $T^{\text{real}}_{\left< s \right>}$ during each inference. Next, it calculates a latency scaling factor $SF$ as the ratio between the actual time and the offline latency estimate up to the split $s$, \textit{i.e.} \mbox{$SF$ = $\frac{T^{\text{real}_{\left< s \right>}}}{T^{\text{offline}_{\left< s \right>}}}$}. 
As a second step, the profiler treats the scaling factor as an indicator of the load of the client device, 
and uses it to estimate the latency of all other candidate splits. 
Thus, the latency of a different split $s'$ is estimated as $SF~\cdot~T^{\text{offline}}_{\left< s' \right>}$. 

Similarly, to assess the server load, the remote endpoint's compute latency is optionally communicated back to the device, piggybacked with the CNN response when offloading. If the server does not communicate back latencies for preserving the privacy of the provider, these can be coarsely estimated as 
$T^{\text{server}}_{\left< s \right>}$ = $T^{\text{response}}_{\left< s,e \right>} - \left(L + \frac{D_{\text{response}}}{B}\right)$, where $T^{\text{response}}_{\left< s \right>}$ is the total time for the server to respond with the result for split point $s$ and exit $e$, $D_{\text{response}}$ is the size of transferred data and $B$, $L$ are the instantaneous network bandwidth and latency respectively.
We periodically offload to the server without stopping the local execution to reassess when the server transitions from ``overloaded'' to ``accepting requests''.


To estimate the instantaneous \textit{network transfer} latency, the profiler employs a run-time monitoring mechanism of the bandwidth $B$ and latency $L$ experienced by the device~\cite{nwslite_2004}. The overall transfer time is
$L + \frac{D_{\left< s \right>}}{B}$ 
, where $D_{\left< s \right>}$ is the amount of data to be transferred given split $s$.
As the network conditions change, the monitoring module refines its estimates by means of two moving averages: a real-time estimation \mbox{($L^{\text{rt}}$, $B^{\text{rt}}$)} and a historical moving average ($L^{\text{hist}}$, $B^{\text{hist}}$). The former is updated and used only when transfers have occurred within the last minutes.
If no such information exists, the historical estimates for the same network type are used.

\setlength{\textfloatsep}{0pt}
\begin{algorithm}[!t]
    \footnotesize
    \SetAlgoLined
    \LinesNumbered
    \DontPrintSemicolon
    \KwIn{Space of candidate designs $\Sigma$}
    \nonl
    \myinput{Prioritised hard constraints $\left<C_1, C_2, ..., C_n\right>$ }
    \nonl
    \myinput{Prioritised soft targets $\left<O_1, O_2, ..., O_{|\mathcal{M}|} \right>$}
    \nonl
    \myinput{Current network conditions $net=\left<L, B\right>$}
    \nonl
    \myinput{Current device and server loads $l^{\{\text{dev}, \text{server}\}}$}
    \nonl
    \myinput{Profiler data $prf$}
    \nonl
    \KwOut{Highest performing design $\sigma^* = \left<s^*, thr_{\text{conf}}^* \right>$}
    \nonl\;
    
    $prf \leftarrow$ \text{UpdateTimings}($prf$, $net$, $l^{\text{dev}}$, $l^{\text{server}}$)\\
    $\Sigma^{\text{feasible}} \leftarrow \Sigma$\\
    
    /* - - - \textit{Obtain feasible space based on hard constraints} - - -*/ \\
    \ForEach{$C_i \in \left<C_1, C_2, ..., C_n\right>$}{
        $\Sigma^{\text{feasible}} \leftarrow$ RemoveInfeasiblePoints($prf$, $C_i$, $\Sigma^{\text{feasible}}$)\\
        \rotatebox[origin=c]{180}{$\Lsh$} \textbf{\texttt{VecCompare(}}$prf$, $\Sigma^{\text{feasible}}$(:,$M_i$), $op_i$, $thr_i$\textbf{\texttt{)}} $~\forall i \in [1,n]$\\
    }
    
    
    /* - - - \textit{Optimise user-defined metrics - Eq. (\ref{eq:moo_lex})} - - - */ \\
    $\sigma^* \leftarrow$ OptimiseUserMetrics($prf$, $\left<O_1, O_2, ..., O_{|\mathcal{M}|}\right>$, $\Sigma^{\text{feasible}}$) \\
    \rotatebox[origin=c]{180}{$\boldsymbol{\Lsh}$} \textbf{\texttt{VecMax/Min(}}$prf$, $\Sigma^{\text{feasible}}$(:,$M_i$), $op_i$\textbf{\texttt{)}} $~\forall i \in [1,|\mathcal{M}|]$
    
    \caption{\footnotesize Flow of dynamic scheduler upon invocation}
    \label{alg:dynamic_sched}
\end{algorithm}

\vspace{-1em}
\subsection{Dynamic Scheduler}
\label{sec:opt_framework}

Given the output of the profiler, the dynamic scheduler (Figure~\ref{fig:dope-nn-arch}~\circled{4}) is responsible for distributing the computation between device and cloud, and deciding the early-exit policy of the progressive inference network. Its goal is to yield the highest performing configuration that satisfies the app requirements.
%
%
To enable the support of realistic multi-criteria SLAs, the scheduler incorporates a combination of \textit{hard constraints} (\textit{e.g.} a strict inference latency deadline of 100 ms) and \textit{soft targets} (\textit{e.g.} minimise cost on the device side). 
Internally, we capture these under a multi-objective optimisation (MOO) formulation. The current set of metrics is defined as
\vspace{-0.15cm}
\begin{small}
    \begin{equation}
        \label{eq:metric_set}
        \mathcal{M} = \{latency, throughput, server~cost, device~cost, accuracy\} \nonumber
        \vspace{-0.15cm}
    \end{equation}
\end{small}
In \tool{}, we interpret cloud and device cost as the execution time on the respective side. 
The defined metrics set $\mathcal{M}$, together with the associated constraints, can cover a wide range of use-cases, based on the relative importance between the metrics for the target task.
Formally, we define a hard constraint as $C=\left<M, op, thr \right>$ where $M \in \mathcal{M}$ is a metric, $op$ is an operator, \textit{e.g.} $\le$, and $thr$  \mbox{is a threshold value}. 

With respect to soft optimisation targets, we define them formally as $O=\left<M, min/max/value \right>$ where a given metric $M \in \mathcal{M}$ is either maximised, minimised or as close as possible to a desirable value. To enable the user to specify the importance of each metric, we employ a multi-objective lexicographic formulation \cite{marler2004survey}, \mbox{shown in \mbox{Eq.~(\ref{eq:moo_lex})}}.
\vspace{-0.15cm}
\begin{align}
    \label{eq:moo_lex}
    &\min\limits_{\sigma} ~M_i(\sigma), ~\text{s.t. }~ M_j(\sigma) \le M_j(\sigma_j^*)~\\ &~j=1,2,...,i-1~,~i>1~,~i=1,2,...,|\mathcal{M}| \nonumber
\end{align}
\vspace{-1.5em}

\noindent
where $\sigma$ represents a design point, $M_i \in \left<M_1, M_2, ...,M_{|\mathcal{M}|} \right>$ is the i-th metric in the ordered tuple of soft targets, $i$ is a metric's position in the importance sequence and $M_j(\sigma_j^*)$ represents the optimum of the j-th metric, found in the j-th iteration. Under this formulation, the user ranks the metrics in order of importance as required by the target use-case.

Algorithm \ref{alg:dynamic_sched} presents the scheduler's processing flow. As a first step, the scheduler uses the estimated network latency and bandwidth, and device and server loads to update the profiler parameters (line 1), including the attainable latency and throughput, and device and server cost for each candidate configuration. 
Next, all infeasible solutions are discarded based on the supplied hard constraints (lines 4-7); given an ordered tuple of prioritised constraints $\left<C_1, C_2, ..., C_n \right>$, the scheduler iteratively eliminates all configurations $\sigma$$=$$\left<s, thr_{\text{conf}}\right>$ that violate them in the given order, where $s$ and $thr_{\text{conf}}$ represent the associated split point and confidence threshold respectively. In case there is no configuration to satisfy all the constraints up to i+1, the scheduler adopts a best-effort strategy by keeping the solutions that comply with up to the i-th constraint and treating the remaining $n$-$i$ constraints as soft targets.
Finally, the scheduler performs a lexicographic optimisation of the user-prioritised soft targets (lines 9-10). To determine the highest performing configuration $\sigma^*$, the scheduler solves a sequence of $|\mathcal{M}|$ single-objective optimisation problems, \textit{i.e.} one for each $M$$\in$$\left<M_1, M_2, ..., M_{|\mathcal{M}|} \right>$~(Eq.~(\ref{eq:moo_lex})).


\noindent

\textbf{Deployment.}
Upon deployment, the scheduler is run on the client side, since most relevant information resides on-device. In a multi-client setting, this setup is further reinforced by the fact that each client device decides independently on its offloading parameters.
%
However, to be deployable without throttling the resources of the target mobile platform, the scheduler has to yield low resource utilisation at run time. To this end, we vectorise the comparison, maximisation and minimisation operations (lines 5-6 and 9-10 in Algorithm \ref{alg:dynamic_sched}) to utilise the SIMD instructions of the target mobile CPU (\textit{e.g.} the NEON instructions on ARM-based cores) and minimise the scheduler's runtime.

At run time, although the overhead of the scheduler is relatively low, \tool{} only re-evaluates the candidate configurations when the outputs of the profiler change by more than a predefined threshold. For highly transient workloads, we can switch from a moving average to an exponential back-off threshold model for mitigating too many scheduler calls. The scheduler overhead and the tuning of the invocation frequency is discussed in Section~\ref{sec:net_variation}. 

The server -- or HA proxy\footnote{High-Availability proxy for load balancing \& fault tolerance in \mbox{data centres}.} \cite{tarreau2012haproxy} in multi-server architectures -- can admit and schedule requests on the remote side to balance the workload and minimise inference latency, maximise throughput or minimise the overall cost by dynamically scaling down unused resources.  We consider these optimisations cloud-specific and out of the scope of this paper. As a result, in our experiments we account for a single server always spinning and having the model resident to its memory. Nevertheless, in a typical deployment, we would envision a caching proxy serving the models with RDMA to the CPU or GPU
of the end server, in a virtualised or serverless environment so as to tackle the cold-start problem~\cite{serverless_oakes2018,serverless_wang2018}.  \blue{Furthermore, to avoid  oscillations (flapping) of computation between the deployed devices and the available servers, techniques used for data-center traffic flapping are employed~\cite{flap}.}

\subsection{CNN Communication Optimiser}
\label{sec:comms_opt}

CNN layers often produce large volumes of intermediate data which come with a high penalty in terms of network transfer.
A key enabler in alleviating the communication bottleneck in \tool{} is the communication optimiser module (\texttt{CNN-CO}) (Figure~\ref{fig:dope-nn-arch}~\circled{5}). \texttt{CNN-CO} comprises two stages.
In the first stage, we exploit the resilience of CNNs to low-precision representation~\cite{guo2017angel,gysel_2018,cascadecnn2018,8_bit_quant_2018} and lower the data precision from 32-bit floating-point down to 8-bit fixed-point through linear quantisation~\cite{8b_tensorrt_2017,8_bit_quant_2018}. By reducing the bitwidth of \textit{only} the data to be transferred, our scheme allows the transfer size to be substantially lower without significant impact on the accuracy of the subsequent classifiers (\textit{i.e.} <0.65 percentage point drop across all exits of the examined CNNs).
Our scheme differs from both i) \textit{weights-only reduction}~\cite{han2016deep,Zhou2017inq,Stock_2020}, which minimises the model size rather than activations' size and ii) \textit{all-layers quantisation}~\cite{qnns_2017,guo2017angel,gysel_2018,8_bit_quant_2018} which requires complex techniques, such as quantisation-aware training~\cite{qnns_2017,8_bit_quant_2018} or a re-training step~\cite{guo2017angel,gysel_2018}, to recover the accuracy drop due to the precision reduction across all layers. 

The second stage exploits the observation that activation data are amenable to compression. A significant fraction of activations are zero-valued, meaning that they are sparse and highly compressible. As noted by prior works~\cite{xu2015empirical,cdma_2018,nikolic2019characterizing}, this sparsity of activations is due to the extensive use of the \texttt{ReLU} layer that follows the majority of layers in modern CNNs. In \texttt{CNN-CO}, sparsity is further magnified due to the reduced precision. In this respect, \texttt{CNN-CO} leverages the sparsity of the 8-bit data by means of an \texttt{LZ4} compressor \mbox{with bit shuffling}.

At run time, \tool{} predicts whether the compression cost will outweigh its benefits by comparing the estimated \texttt{CNN-CO} runtime to the transfer time savings. 
If \texttt{CNN-CO}'s overhead is amortised, \tool{} queues offloading requests' data to the \texttt{CNN-CO}, with dedicated threads for each of the two stages, before transmission. Upon reception at the remote end, the data are decompressed and cast back to the original precision to continue inference. The overhead is shown as \textit{packing} in Figure~\ref{fig:jetson-times} for non-progressive models.



\subsection{Distributed Execution Engine} 
\label{sec:offload}

In popular Deep Learning frameworks, such as \textit{TensorFlow}~\cite{tensorflow_2016} and \textit{PyTorch}~\cite{pytorch2019}, layers are represented by \emph{modules} and data in the form of multi-dimensional matrices, called \emph{tensors}.
To split and offload computation, \tool{} modifies CNN layer's operations behind the scenes.
To achieve this, it intercepts module and tensor operations by replacing their functions with a custom wrapper using Python's function decorators. 





\begin{figure}[t]
  \centering
  \vspace{-0.15cm}
  \includegraphics[trim={0.5cm 9cm 7.5cm 0cm},clip,width=0.49\textwidth]{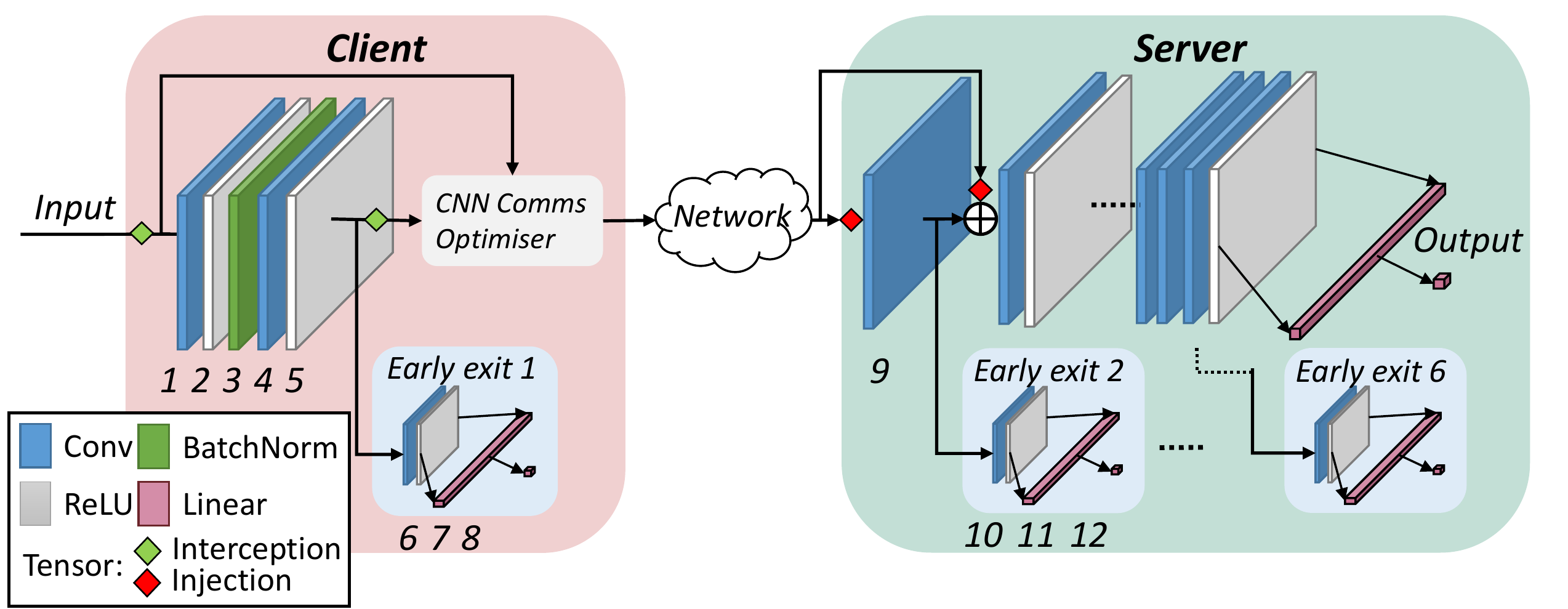}
  \vspace{-0.65cm}
  \caption{Offloading a progressive ResNet block.}
  \label{fig:spinn-system}
  \vspace{0.1cm}
\end{figure}

Figure \ref{fig:spinn-system} focuses on an instance of an example ResNet block. \tool{} attributes IDs to each layer in a trace-based manner, by executing the network and using the layer's execution order as a sequence identifier.\footnote{Despite the existence of branches, CNN execution tends to be parallelised across data rather than layers. Hence, the numbering is deterministic.} 
\tool{} uses these IDs to build an execution graph in the form of a directed acyclic graph (DAG), with nodes representing tensor operations and edges the tensor flows among them~\cite{tensorflow_2016,Sharma_2016, fpgaconvnet_2018}. This is then used to detect the dependencies across partitions. To achieve this, Python's dynamic instance attribute registration is used to \texttt{taint} tensors and monitor their flow through the network.
With Figure \ref{fig:spinn-system} as a reference, \tool{}'s custom wrapper (Figure~\ref{fig:dope-nn-arch}~\circled{6}) performs the following operations:

\textbf{Normal execution:} When a layer is to be run locally, the wrapper calls the original function it replaced.
In Figure~\ref{fig:spinn-system}, layers 1 to 8 execute normally on-device, while layers from 9 until the end execute normally on the server side.
    
\textbf{Offload execution:} When a layer is selected as a partition point (layer 9), instead of executing the original computation on the client, the wrapper queues an offloading request to be transmitted. This request contains the inputs (layer 5's output) and subsequent layer dependencies (input of skip connection). Furthermore, the inference and the data transfer are decoupled in two separate threads, to allow for pipelining of data transfer and next frame processing.
    
\textbf{Resume execution:} Upon receiving an offloading request, the inputs and dependencies are injected in the respective layers (layer 9 and \textit{add} operation) and normal execution is resumed on the remote side. When the server concludes execution, the results are sent back in a parallel thread.
    
\textbf{Early exit:} When an intermediate classifier (\textit{i.e.} early exit) is executed, the wrapper evaluates its prediction confidence (Eq. (\ref{eq:softmax})). If it is above the provided ${thr}_{\text{conf}}$, execution terminates early, returning the current prediction (Eq.~(\ref{eq:conf_threshold})). 
    
%
%


Since the premise of our system is to always have at least one usable result on the client side, we continue the computation on-device even past the split layer, until the next early exit is encountered.
Furthermore, to avoid wasteful data transmission and redundant computation, if a client-side early exit covers the latency SLA and satisfies the selected $thr_{\text{conf}}$, the client sends a termination signal to the remote worker to cancel the rest of the inference. If remote execution has not started yet, \tool{} does not offload at all.




%% file: sections/evaluation.tex
\vspace{-0.5em}
\section{Evaluation}
\label{sec:eval}

This section presents the effectiveness of \tool{} in significantly improving the performance of mobile CNN inference by examining its core components and comparing with the currently standard device- and cloud-only implementations and state-of-the-art collaborative inference systems.

\vspace{-1em}
\subsection{Experimental Setup}
\label{sec:exp_setup}


\setlength{\tabcolsep}{2pt}
\begin{table}[t]
    \centering
    \scriptsize
    \resizebox{\linewidth}{!}{
    \begin{tabu}{l l c c l l l l} 
        \toprule
        \begin{tabular}{@{}c@{}}\textbf{Platform} \\  \end{tabular} & \begin{tabular}{@{}c@{}} \textbf{CPU} \\  \end{tabular} 
        & \begin{tabular}{@{}c@{}} \textbf{Clock Freq.} \\ \end{tabular}
        & \begin{tabular}{@{}c@{}} \textbf{Memory} \\  \end{tabular} 
         & \begin{tabular}{@{}l@{}}  \textbf{GPU}  \\  \end{tabular}
        \\ \midrule
        Server 
        & 2$\times$ Intel Xeon Gold 6130 
        & 2.10 GHz
        & 128 GB
        & GTX1080Ti 
        \\ 
        Jetson AGX 
        & Carmel ARMv8.2 
        & 2.26 GHz
        & \phantom{1}16 GB
        & 512-core Volta 
        \\
        \bottomrule
    \end{tabu}
    }
    \vspace{0.05cm}
    \caption{Specifications of evaluated platforms.
    }
    \vspace{-0.25cm}
    \label{tab:experimental_setup}
\end{table}

For our experiments, we used a powerful computer as the server and an Nvidia Jetson Xavier AGX as the client (Table~\ref{tab:experimental_setup}). 
Specifically for Jetson,  we tested against three different power profiles to emulate end-devices with different compute capabilities:\footnote{We are adjusting the TDP and clock frequency of the CPU and GPU cores, \blue{effectively emulating different tiers of devices, ranging from high-end embedded devices to mid-tier smartphones.}} 1) \textit{30W} (full power), 2) \textit{10W} (low power), \mbox{3) \textit{underclocked 10W}} (u10W). 
Furthermore, to study the effect of limited computation capacity (\textit{e.g.} high-load server), we emulated the load by linearly scaling up the CNN computation times on the server side. 
We simulated the network conditions of offloading by using the average bandwidth and latency across national carriers \cite{4g_speed,5g_speed},
for 3G, 4G and 5G mobile networks. For local-area connections (Gigabit Ethernet 802.3, WiFi-5 802.11ac), we used the nominal speeds of the protocol. 
We have developed \tool{} on top of \textit{PyTorch} (1.1.0) and experimented with four models, altered from \textit{torchvision} (0.3.0) to include early exits or to reflect the CIFAR-specific architectural changes. We evaluated \tool{} using: ResNet-50 and -56~\cite{He_2016}, VGG16~\cite{Simonyan14c}, MobileNetV2~\cite{Sandler} and \mbox{Inception-v3}~\cite{Szegedy2015}.
Unless stated otherwise, each benchmark was run 50 times to obtain the average latency. 

\noindent
\textbf{Datasets and Training.}
We evaluated \tool{} on two datasets, namely CIFAR-100~\cite{Krizhevsky2009} and ImageNet (ILSVRC2012)~\cite{Fei-Fei2010}. The former contains 50k training and 10k test images of resolution $32$$\times$$32$, each corresponding to one of 100 labels.
The latter is significantly larger, with 1.2m training and 50k test images of $300$$\times$$300$ and 1000 labels.
We used the preprocessing steps described in each model's implementation, such as \textit{scaling} and \textit{cropping} the input image, stochastic image \textit{flip} ($p=0.5$) and colour channel \textit{normalisation}.
After converting these models to progressive early-exit networks, we trained them jointly from scratch end-to-end, with the ``overthink" cost function (Section \ref{sec:prog_inf}). 
We used the authors' training hyperparameters, except for MobileNetV2, where we utilised SGD with learning rate of 0.05 and cosine learning rate scheduling, due to convergence issues. 
We trained the networks for 300 epochs on CIFAR-100 and 90 epochs on ImageNet. 

\begin{figure}[t]
    \centering
    \includegraphics[width=0.97\columnwidth]{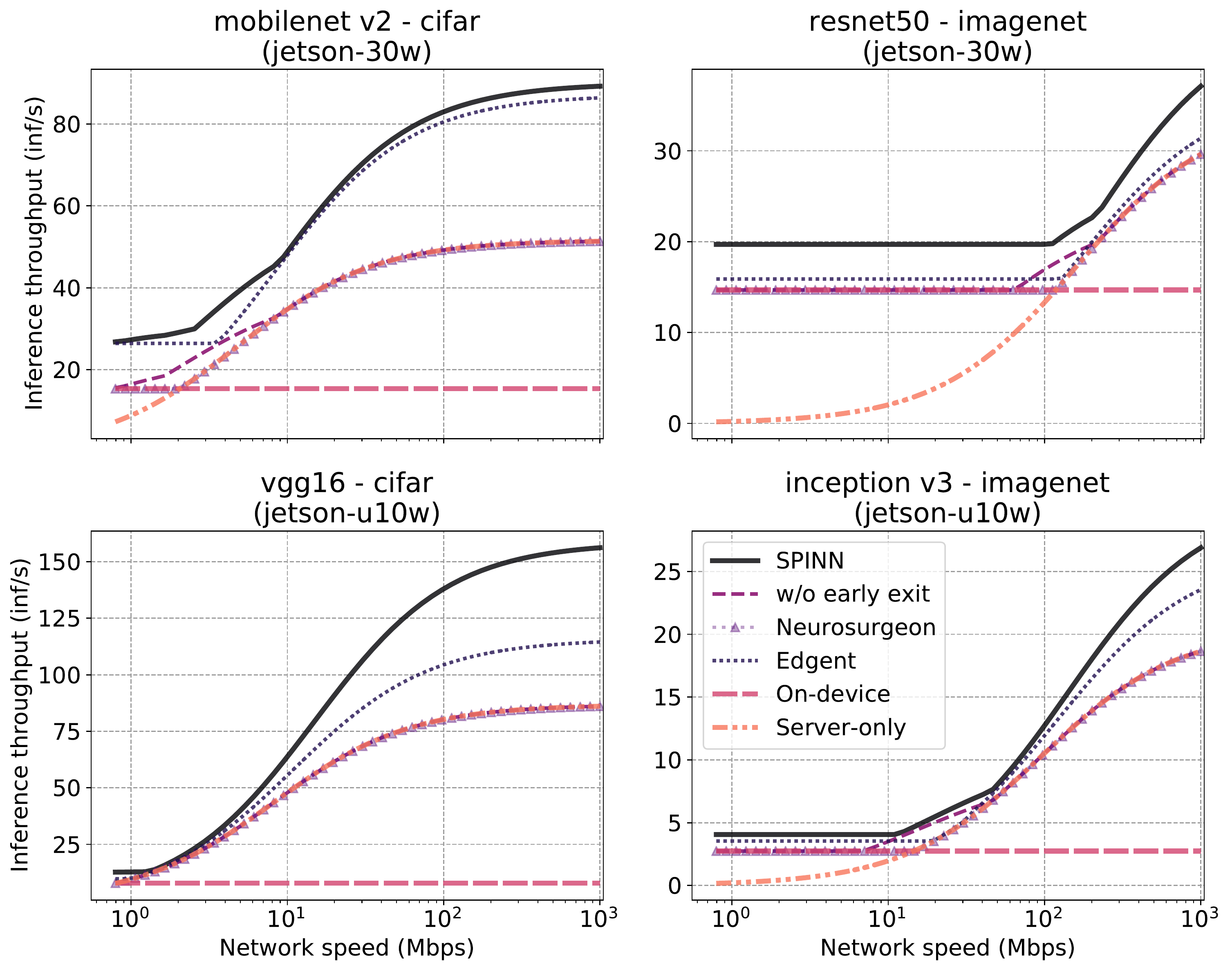}
    \vspace{-0.3cm}
    \caption{Achieved throughput for various $\left<\text{model, device, dataset}\right>$ setups vs. network speed.}
    \label{fig:thrpt_comparison}
    \vspace{0.15cm}
\end{figure}
\vspace{-0.25cm}
\subsection{Performance Comparison}
\label{sec:perf_comparison}
This subsection presents a performance comparison of \tool{} with: 1)~the two state-of-the-art CNN offloading systems \texttt{Neurosurgeon} \cite{Kang2017} and \texttt{Edgent}~\cite{edgent_2020} (Section~\ref{sec:background}); 2)~the status-quo cloud- and device-only baselines; and 3)~a non-progressive ablated variant of \tool{}. 

\vspace{-0.15cm}
\subsubsection{Throughput Maximisation}
Here, we assess \tool{}'s inference throughput across varying network conditions. 
For these experiments, the \tool{} scheduler’s objectives were set to maximise throughput with up to 1 percentage \mbox{point (pp)} tolerance in accuracy drop with respect to the CNN's \mbox{last exit}.

Figure \ref{fig:thrpt_comparison} shows the achieved inference throughput for varying  network speeds.
On-device execution yields the same throughput independently of the network variation, but is constrained by the processing power of the client device. Server-only execution follows the trajectory of the available bandwidth.
\texttt{Edgent} always executes a part of the CNN (up to the exit that satisfies the 1pp accuracy tolerance) irrespective of the network conditions.  As a result, it follows a similar trajectory to server-only but achieves higher throughput due to executing only part of the model. 
\texttt{Neurosurgeon} demonstrates a more polarised behaviour; under constrained connectivity it executes the whole model on-device, whereas as bandwidth increases it switches to offloading all computation as it results in higher throughput.
The ablated variant of \tool{} (\textit{i.e.} without early exits) largely follows the behaviour of \texttt{Neurosurgeon} at the two extremes of the bandwidth while in the middle range, it is able to achieve higher throughput by offloading earlier due to \texttt{CNN-CO} compressing the transferred data.

The end-to-end performance achieved by \tool{} delivers the highest throughput across all setups, achieving a speedup of up to 83\% and 52\%
over \texttt{Neurosurgeon} and \texttt{Edgent}, respectively.
This can be attributed to our bandwidth- and data-locality-aware scheduler choices on the early-exit policy and partition point. 
In low bandwidths, \tool{} selects device-only execution, outperforming all other on-device designs due to its early-exiting mechanism, tuned by the scheduler module. In the mid-range, the \texttt{CNN-CO} module enables \tool{} to better utilise the available bandwidth and start offloading earlier on, outperforming both \texttt{Edgent} and \texttt{Neurosurgeon}.
In high-bandwidth settings, our system surpasses the performance of all other designs by exploiting its optimised early-exiting scheme.

\begin{figure}[t]
    \centering
    \includegraphics[width=.99\columnwidth]{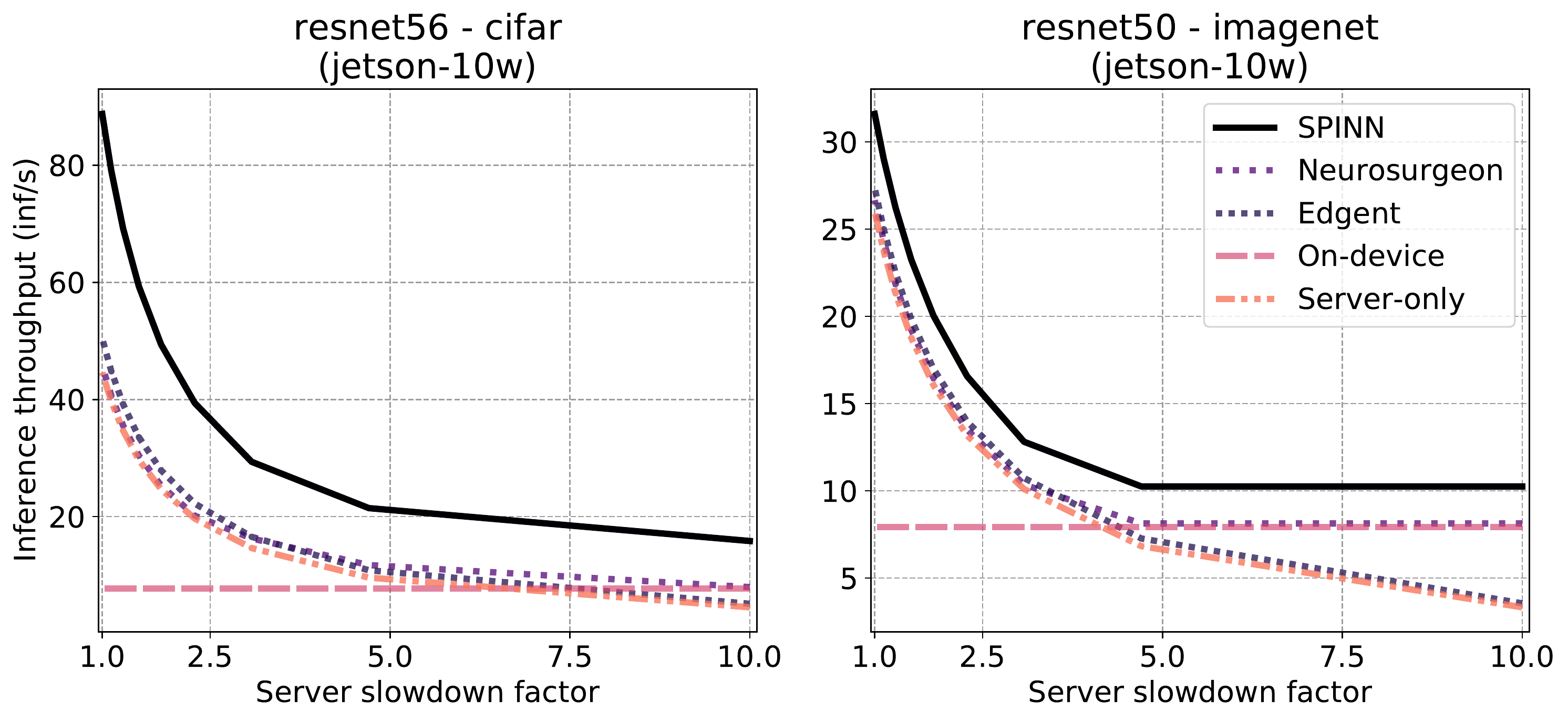}
    \vspace{-0.3cm}
    \caption{Effect of server slowdown on ResNet.}
    \label{fig:server_slowdown}
    \vspace{0.1cm}
\end{figure}

Specifically, compared to \texttt{Edgent}, \tool{} takes advantage of the input's classification difficulty to exit early, whereas the latter only selects an intermediate exit to \textit{uniformly} classify all incoming samples. Moreover, in contrast with \texttt{Edgent}'s strategy to always transmit the input to the remote endpoint, we exploit the fact that data already reside on the device and avoid the wasteful data transfers.

\vspace{-0.15cm}
\subsubsection{Server-Load Variation}
\label{sec:server_load}

To investigate the performance of \tool{} under various server-side loads, we measured the inference throughput of \tool{} against baselines when varying the load of the remote end, with 1pp of accuracy drop tolerance. This is accomplished by linearly scaling the latency of the server execution by a slowdown factor (\textit{i.e.} a factor of 2 means the server is 2$\times$ slower). 
Figure~\ref{fig:server_slowdown} presents the throughput achieved by each approach under various server-side loads, with the Jetson configured at the 10W profile and the network speed in the WiFi-5 range (500 Mbps).


With low server load (left of the x-axis), the examined systems demonstrate a similar trend to the high-bandwidth performance of Figure~\ref{fig:thrpt_comparison}.
As the server becomes more loaded (\textit{i.e.} towards the right-hand side), performance deteriorates, except for the case of device-only execution which is invariant to the server load.
On the one hand, although its attainable throughput reduces, \texttt{Neurosurgeon} adapts its policy based on the server utilisation and gradually executes a greater fraction of the CNN on the client side.
On the other hand, \texttt{Edgent}'s throughput deteriorates more rapidly and even reaches below the device-only execution under high server load, since its run-time mechanism does not consider the varying server load.
Instead, by adaptively optimising both the split point and the early-exit policy, \tool{}'s scheduler manages to adapt the overall execution based on the server-side load, leading to throughput gains between \mbox{1.18-1.99$\times$} (1.57$\times$ geo. mean) and 1.15-3.09$\times$ (1.61$\times$ geo. mean) over \texttt{Neurosurgeon} and \texttt{Edgent} respectively.
%
%
%
\vspace{-0.15cm}
\subsubsection{Case Study: Latency-driven SLAs at minimal server cost}
\label{sec:sla_deadlines}

\begin{figure}[t]
    \centering
    \includegraphics[width=0.44\textwidth]{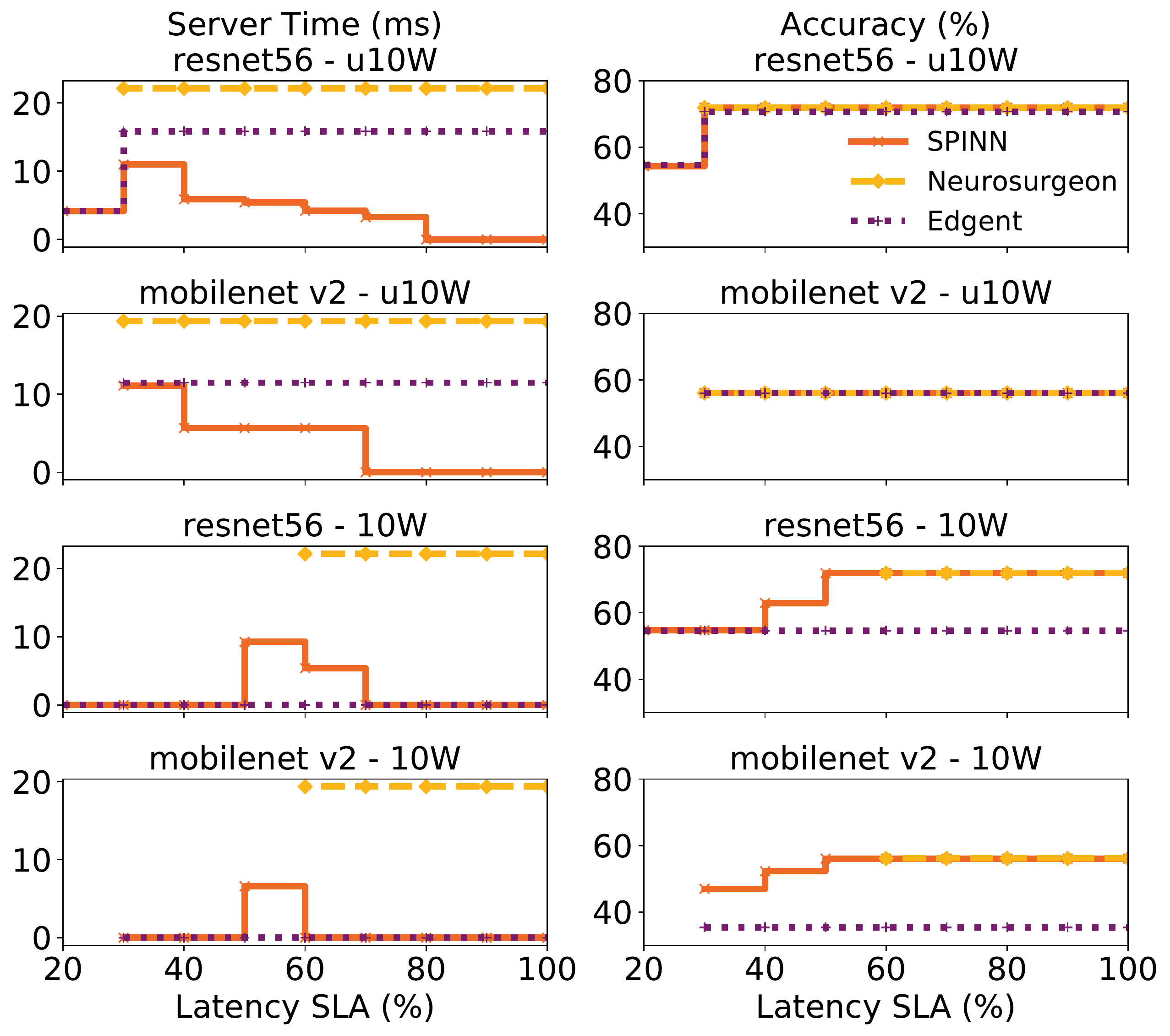}
    \vspace{-0.4cm}
    \caption{Server time (left) and accuracy (right) of \tool{} vs. \texttt{Neurosurgeon} and \texttt{Edgent} for different latency SLAs  and client compute power (u10W and 10W). The SLA is expressed as a percentage of the on-device latency.}
    \label{fig:slas}
    \vspace{0.2cm}
\end{figure}

To assess \tool{}'s performance under deadlines, we target the scenario where a service provider aims to deploy a CNN-based application that meets strict latency SLAs at maximum accuracy and minimal server-side cost. 
In this setting, we compare against \texttt{Neurosurgeon}\footnote{\blue{It should be noted that Neurosurgeon maintains the accuracy of the original CNN.}} and \texttt{Edgent}, targeting MobileNetV2 and ResNet-56 over 4G.
Figure \ref{fig:slas} shows the server computation time and accuracy achieved by each system for two device profiles with different compute capabilities - u10W and 10W. Latency SLAs are represented as a percentage of the device-only runtime of the original CNN (\textit{i.e.} 20\% SLA means that the target is $5\times$ less latency than on-device execution, requiring early exiting and/or server support).

For the low-end device (u10W) and strict latency deadlines, \tool{} offloads as much as possible to the cloud as it allows reaching faster a later exit in the network, hence increasing the accuracy.
As the SLA loosens (reaching more than 40\% of the on-device latency), \tool{} starts to gradually execute more and more locally.
In contrast, \texttt{Edgent} and \texttt{Neurosurgeon} achieve similar accuracy but with up to 4.9$\times$ and 6.8$\times$ higher server load. On average across all targets, \tool{} reduces \texttt{Edgent} and \texttt{Neurosurgeon} server times by 68.64\% and 82.5\% (60.3\% and 83.6\% geo. mean), respectively, due to its flexible multi-objective scheduler.
Instead, \texttt{Neurosurgeon} can only optimise for overall latency 
and cannot trade off accuracy to meet the deadline (\textit{e.g.}~for 20\% SLA on ResNet-56) while \texttt{Edgent} cannot account for server-time minimisation and accuracy drop constraints.

The situation is different for the more powerful device (10W). With the device being faster, the SLA targets become much stricter. Therefore, we observe that \tool{} and \texttt{Edgent} can still meet a latency constraint as low as 20\% and 30\% of the local execution time for ResNet-56 and MobileNetV2 respectively.
In contrast, without progressive inference, it is impossible for \texttt{Neurosurgeon} to achieve inference latency below 60\% of  on-device execution across both CNNs. 
In this context, \tool{} is able to trade off accuracy in order to meet stricter SLAs, but also improve its attainable accuracy as the latency constraints are relaxed.


For looser latency deadlines (target larger than 50\% of the on-device latency), \tool{} achieves accuracy gains of 17.3\% and 20.7\% over \texttt{Edgent} for ResNet-56 and MobileNetV2, respectively.
%
The reason behind this is twofold.
First, when offloading, \texttt{Edgent} starts the computation on the server side, increasing the communication latency overhead. Instead, \tool{}'s client-to-server offloading strategy and compression significantly reduces the communication latency overhead. 
Second, due to \texttt{Edgent}'s unnormalised cost function (\textit{i.e.}~$\max \left(\frac{1}{lat} + acc\right)$), the throughput's reward dominates the accuracy gain, leading to always selecting the first early-exit sub-network and executing it locally.
In contrast, \tool{}'s scheduler's multi-criteria design is able to capture accuracy, server time and latency constraints to  yield an optimised deployment.
Hence, similarly to the slower device, \tool{} successfully exploits the server resources to boost accuracy under latency constraints, while it can reach up to pure on-device execution for loose deadlines.


\begin{figure}[t]
  \centering
  \vspace{-0.15cm}
  \includegraphics[width=0.35\textwidth]{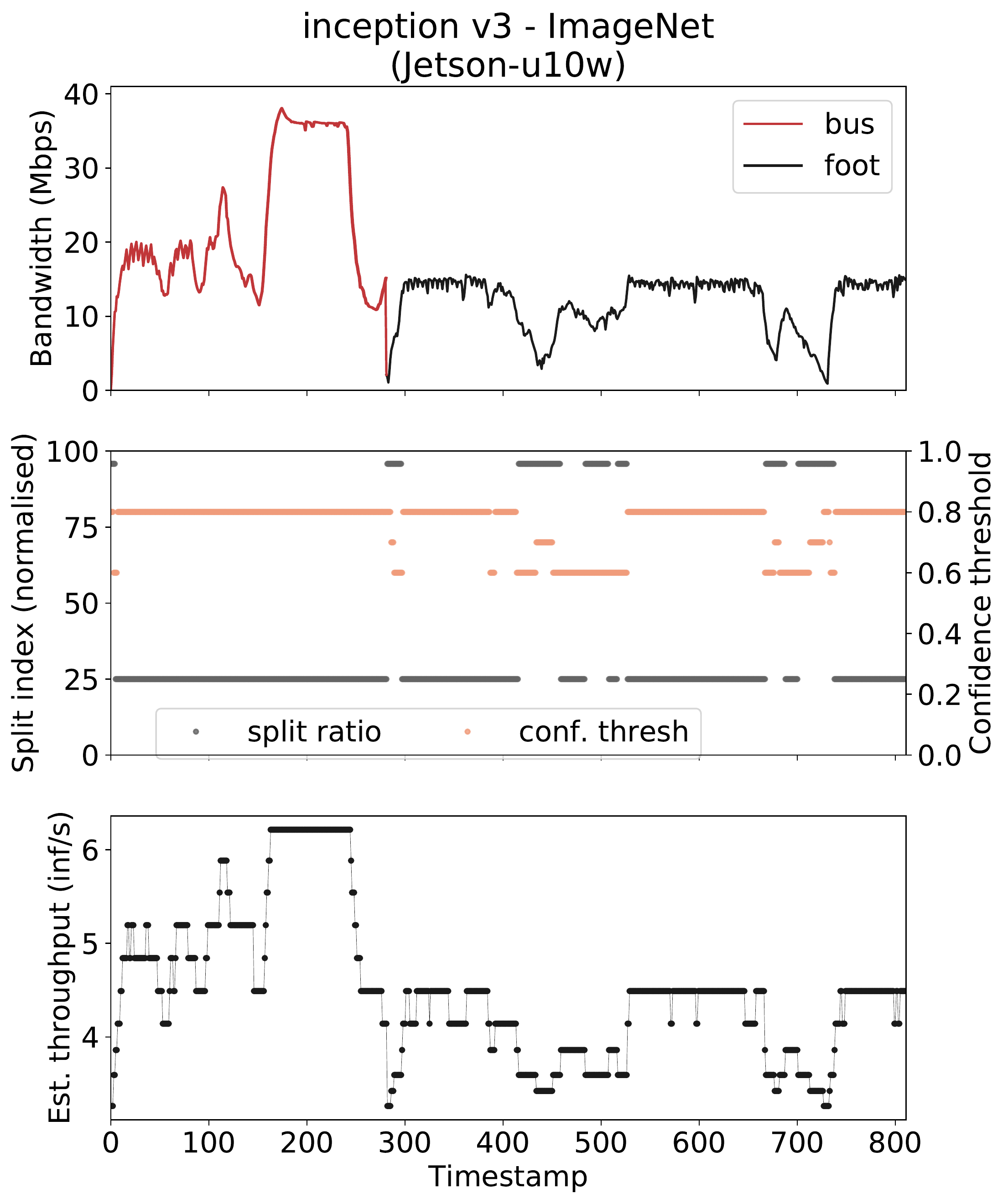}
  \vspace{-0.4cm}
  \caption{SPINN scheduler's behaviour on real \mbox{network} provider trace.}
  \vspace{0.1cm}
  \label{fig:spinn-bw-trace}
\end{figure}

\vspace{-1em}
\subsection{Runtime Overhead and Efficiency}
\label{sec:eval_sched}

\subsubsection{Deployment Overhead}
\label{sec:spinn_overhead}

By evaluating across our examined CNNs and datasets on the CPU of Jetson, the scheduler executes in max 14 ms (11~ms geo. mean). This time includes the cost of reading the profiler parameters, updating the monitored metrics, and searching for and returning the selected configuration.
Moreover, \tool{}'s memory consumption is in the order of a few KB (\textit{i.e.} <1\% of Jetson's RAM). 
These costs are amortised over multiple inferences, as the scheduler is invoked only on significant context changes. We discuss the selection of such parameters in the following section.

\subsubsection{Network Variation}
\label{sec:net_variation}

To assess the responsiveness of \tool{} in adapting to dynamic network conditions, we targeted a real bandwidth trace from a Belgian ISP. The trace contains time series of network bandwidth variability during different user activities.
In this setup, \tool{} executes the ImageNet-trained Inception-v3 with Jetson-u10W as the client under the varying bandwidth emulated by the Belgium 4G/LTE logs. The scheduler is configured to maximise both throughput and accuracy.
%
Figure~\ref{fig:spinn-bw-trace}~(top) shows an example bandwidth trace from a moving bus followed by walking. Figure~\ref{fig:spinn-bw-trace}~(bottom) shows \tool{}'s achieved inference throughput under the changing network quality. The associated scheduler decisions are depicted in \mbox{Figure~\ref{fig:spinn-bw-trace}~(middle)}.

At low bandwidth (<5 Mbps), \tool{} falls back to device-only execution. In these cases, the scheduler adopts a less conservative early-exit policy by lowering the confidence threshold. In this manner, it allows more samples to exit earlier, compensating for the client's low processing power. Nonetheless, the impact on accuracy remains minimal (<$1\%$) for the selected early-exit policies by the scheduler ($thr_\text{conf}~$$\in$$~[0.6, 1.0]$), as illustrated in Figure~\ref{fig:accuracy_conf_theshold} for Inception-v3 on ImageNet.
At the other end, high bandwidths result in selecting an earlier split point and thus achieving up to $7\times$ more inf/sec over pure on-device execution.
Finally, the similar trajectories of the top and bottom figure suggest that our scheduler can adapt the system to the running conditions, without having to be continuously invoked.

Overall, we observe that small bandwidth changes do not cause significant alterations to the split and early-exit strategies. By employing an averaging historical window of three values and a difference threshold of 5\%, the scheduler is invoked $1/3$ of the total bandwidth changes across traces.


\begin{figure}[t]
  \vspace{-1em}
  \centering
  \includegraphics[width=0.42\textwidth]{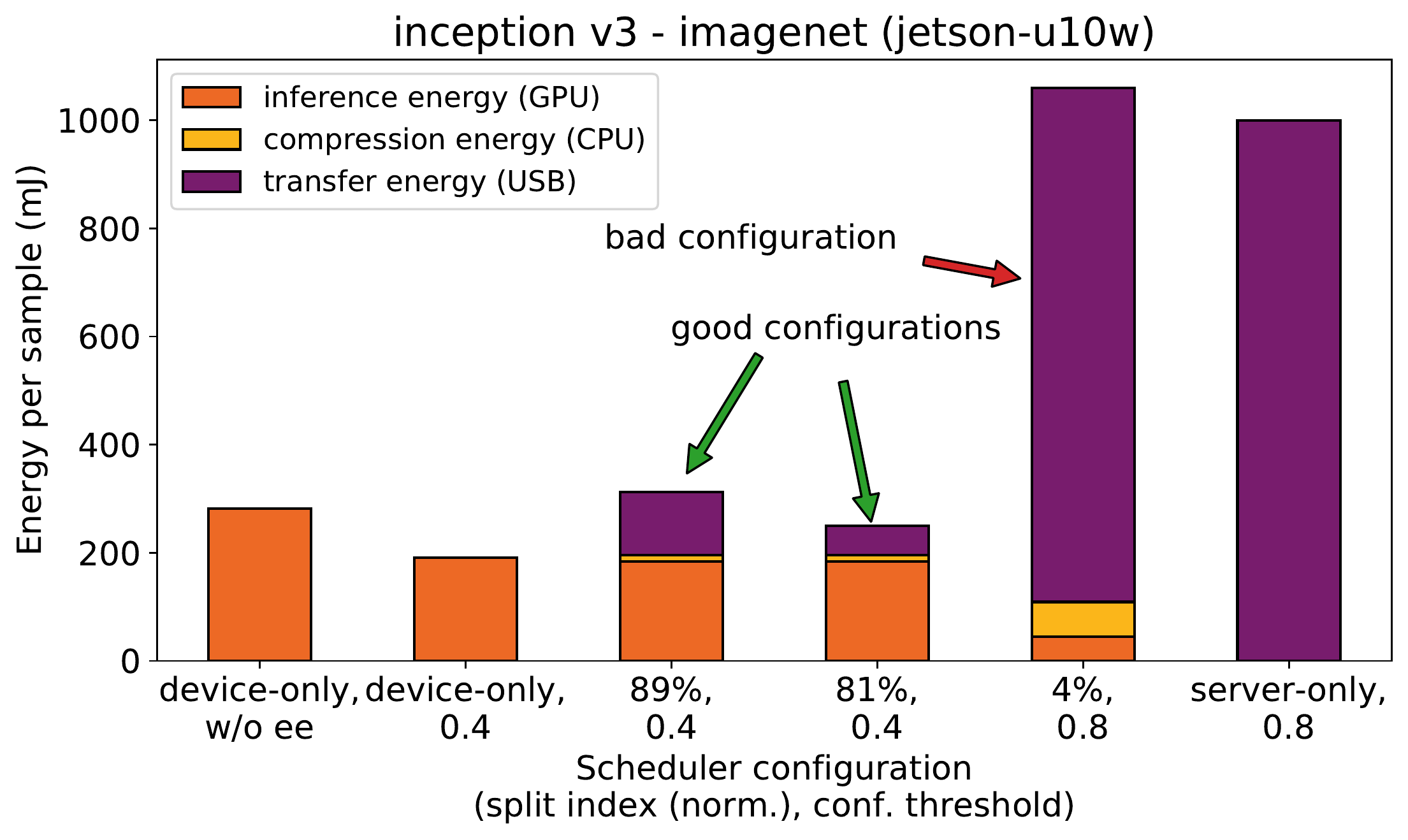}
  \vspace{-0.5cm}
  \caption{Energy consumption of \tool{} vs. baselines.
  \vspace{0.1cm}
  }
  \label{fig:energy}
\end{figure}

\subsection{Energy Consumption}
\label{sec:energy}

Figure~\ref{fig:energy} shows the breakdown of dominant energy consumption across the client device subsystems. We measured energy consumption over 1000 inferences from the validation set and offloading over UK's Three Broadband's 4G network with a Huawei E3372 USB adapter. We measured the power of Jetson (CPU, GPU) from its integrated probes and the transmission energy with the Monsoon AAA10F power monitor.

Traversing the horizontal axis left-to-right, we first see device-only execution without and with early-exits, where the local processing dominates the total energy consumption. The latter shows benefits due to samples exiting early from the network.
Next, we showcase the consumption breakdown of three different $\left< split, thr_\text{conf} \right>$ configurations. The first two configurations demonstrate comparable energy consumption with the device-only execution without early exits. On the contrary, a bad configuration requires an excessively large transfer size, leading to large compression and transfer energy overheads. Last, the energy consumption when fully offloading is dominated by the network transfer.

Across configurations, we witness a $5\times$ difference in energy consumption across different inference setups. While device-only execution yields the lowest energy footprint per sample, it is also the slowest.
Our scheduler is able to yield deployments that are significantly more energy efficient than full offloading ($4.2\times$) and on par with on-device execution ($0.76-1.12\times$), while delivering significantly faster end-to-end processing. Finally, with different configurations varying both in energy and performance, the decision space is amenable to energy-driven optimisation by adding energy as a scheduler optimisation metric.

\vspace{-0.25cm}
    \subsection{Constrained Availability Robustness}
\label{sec:robust}

Next we evaluate \tool{}'s robustness under constrained availability of the remote end such as network timeouts, disconnections and server failures. More specifically, we investigate 1) the achieved accuracy across various failure rates and 2) the latency improvement over conventional systems enhanced with an error-control policy. 
In these experiments, we fix the confidence threshold of three models (Inception-v3, ResNet-56 and ResNet-50) to a value of 0.8 and emulate variable failure rates by sampling from a random distribution across the validation set.

\begin{figure}[t]
    \centering
    \vspace{-1em}
    \includegraphics[width=.96\columnwidth]{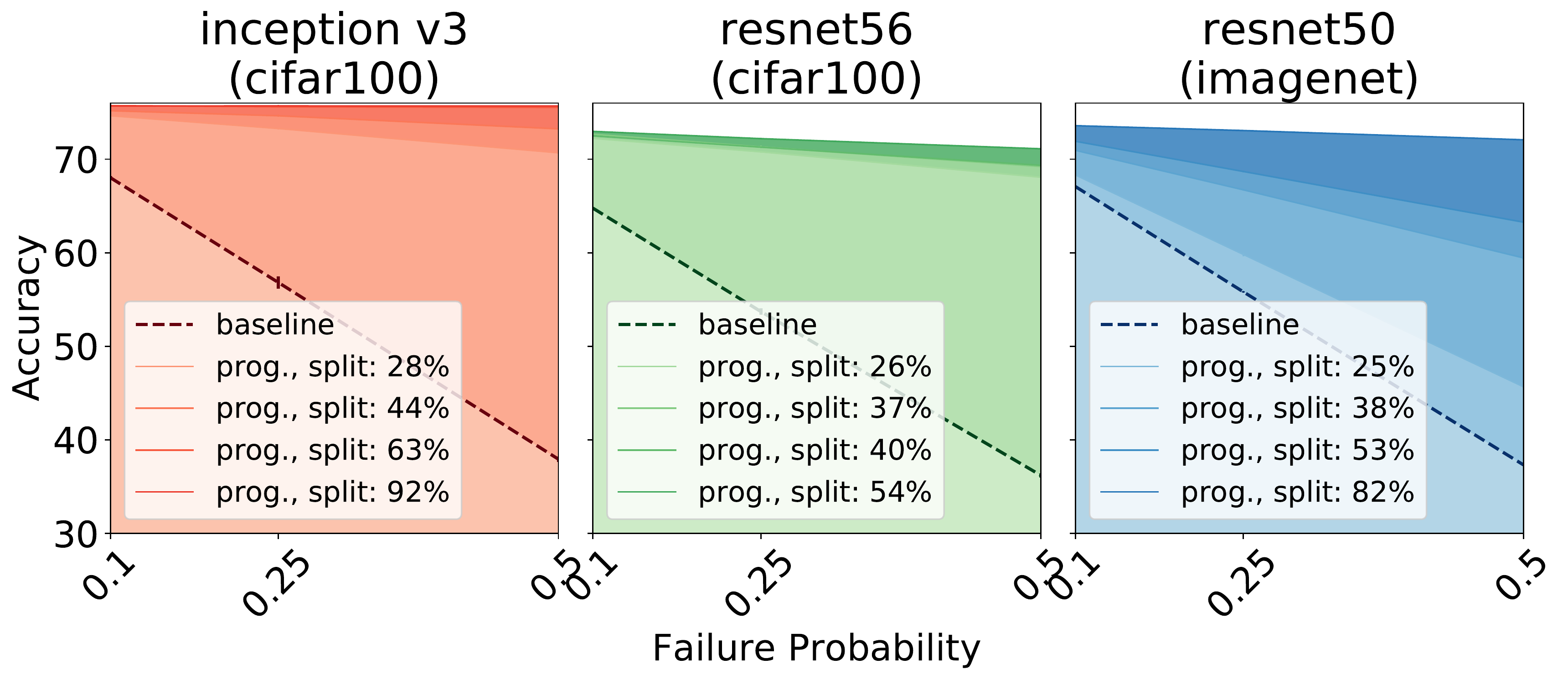}
    \vspace{-0.4cm}
    \caption{Comparison of the average accuracy under uncertain server availability. \blue{The shaded area indicates attained accuracies under a probability distribution.}}
    \label{fig:robustness_acc}
\end{figure}

\textbf{Accuracy comparison}:
For this experiment (Figure \ref{fig:robustness_acc}), we compare \tool{} at different network split points (solid colours) against a non-progressive baseline (dashed line). Under failure conditions, the baseline unavoidably misclassifies the result as there is no usable result locally on-device. However, \tool{} makes it possible to acquire the most confident local result up to the split point, when the server is unavailable. 

As shown in Figure \ref{fig:robustness_acc}, the baseline quickly drops in accuracy as the failure rate increases. This is not the case with \tool{}, which manages to maintain a minimal accuracy drop. Specifically, we witness drops ranging in $[0,5.75\%]$ for CIFAR-100 and $[0.46\%,33\%]$ for ImageΝet, when the equivalent drop of the baseline is $[11.56\%,44.1\%]$ and $[9.25\%,44.34\%]$, respectively. 
As expected, faster devices are able to execute locally a larger part of the model (exit later) while meeting their SLA exhibit the smallest impact under failure, as depicted in the progressive variants of the two models.


\textbf{Latency comparison}:
In this scenario, we compare \tool{} against a single-exit offloaded variant of the same networks. This time instead of simply failing the inference when the remote end is unavailable, we allow for \textit{retransmission} with \textit{exponential back-off}, a common behaviour of distributed systems to avoid channel contention. When a sample fails under the respective probability distribution, the result gets retransmitted. If the same sample fails again, the client waits double the time and retransmits, until it succeeds.
Here, we assume Jetson-10W offloads to our server over 4G and varying failure probability ($P_\text{fail} 
\in \{0.1, 0.25, 0.5\}$). The initial retransmission latency is 20 ms. We ran each experiment three times and report the mean and standard deviation of the latencies.

\begin{figure}[t]
    \centering
    \vspace{-1em}
    \includegraphics[width=.96\columnwidth]{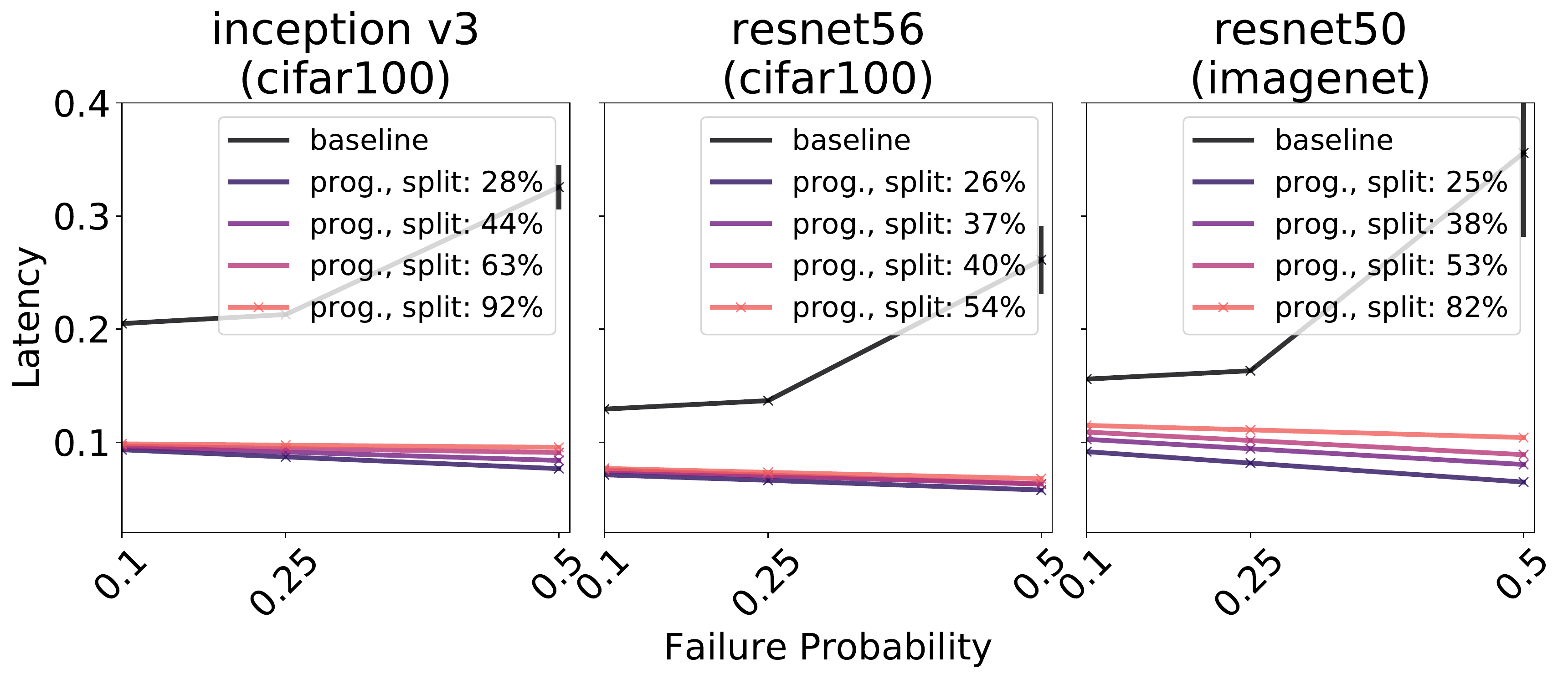}
    \vspace{-0.4cm}
    \caption{Comparison of average latency under \mbox{uncertain} server availability.}
    \label{fig:robustness_latencies}
    \vspace{0.2cm}
\end{figure}

As depicted in Figure \ref{fig:robustness_latencies}, the non-progressive baseline follows a trajectory of increasing latency as the failure probability gets higher, due to the additional back-off latency each time a sample fails. While the impact on the average latency for both networks going from $P_{\text{fail}}=0.1$ to $P_{\text{fail}}=0.25$ is gradual, at 3.9\%, 5.8\% and 4.7\% for {Inception-v3}, {ResNet-56} and {ResNet-50} respectively, the jump from $P_{\text{fail}}=0.25$ to $P_{\text{fail}}=0.5$ is much more dramatic, at 52.9\%, 91\% and 118\%. The variance at $P_{\text{fail}}=0.5$ is also noticeably higher, compared to previous values, attributed to higher number of retransmissions and thus higher discrepancies across different runs.
We should note that despite the considerably higher latency of the non-progressive baseline, its accuracy can be higher, since all samples -- whether re-transmitted or not -- exit at the final classifier.
Last, we also notice a slight reduction in the average latency of \tool{}'s models as $P_{\text{fail}}$ increases. This is a result of more samples early-exiting in the network, as the server becomes unavailable more often.

To sum up, these two results demonstrate that \tool{} can perform sufficiently, in terms of accuracy and latency, even when the remote end remains unresponsive, by falling back to results of local exits. Compared to other systems, as the probability of failure when offloading to the server increases, there is a gradual degradation of the quality of service, instead of catastrophic unresponsiveness of the application.

%% file: sections/limitations.tex
\section{Discussion}

\blue{
\textbf{\tool{} and the current ML landscape.}
The status-quo deployment process of CNNs encompasses the maintenance of two models: a large, highly accurate model on the cloud and a compact, lower-accuracy one on the device. However, this approach comes with significant deployment overheads. First, from a development time perspective, the two-model approach results in two time- and resource-expensive stages. In the first stage, the large model is designed and trained requiring multiple GPU-hours. In the second stage, the large model is compressed through various techniques in order to obtain its lightweight counterpart, with the selection and tuning of the compression method being a difficult task in itself. Furthermore, typically, to gain back the accuracy loss due to compression, the lightweight model has to be fine-tuned through a number of additional training steps.}

\blue{
With regards to the lightweight compressed networks, \tool{} is orthogonal to these techniques and hence a compressed model can be combined with \tool{} to obtain further gains. Given a compressed model, our system would proceed to derive a progressive inference variant with early exits and deploy the network with a tailored implementation. For use-cases where pre-trained models are employed, \tool{} can also smoothly be adopted by modifying its training scheme (Section~\ref{sec:prog_inf}) so that the pre-trained backbone is frozen during training and only the early exits are updated.
}

\blue{Nonetheless, with \tool{} we also enable an alternative paradigm that alleviates the main limitations of the current practice. \tool{} requires a single network design step and a single training process - which trains both the backbone network and its early exits. Upon deployment, the model's execution is adaptively tuned based on the multiple target objectives, the environmental conditions and the device and cloud load. In this manner, \tool{} enables a highly customised deployment which is dynamically and efficiently adjusted to sustain its performance in mobile settings. This approach is further supported by the ML community's growing number of works on progressive networks~\cite{branchynet2016,Huang2017,sdn_icml_2019,li2019improved,Zhang_2019_ICCV,scan2019neurips,deebert2020acl} which can be directly targeted by \tool{} to yield an optimised deployment on mobile platforms.
}


\blue{\textbf{Limitations and future work.}}
Despite the challenges addressed by \tool{}, our prototype system has certain limitations.
First, the scheduler does not explicitly optimise for energy or memory consumption of the client. The energy consumption could be integrated as another objective in the MOO solver of the scheduler, while memory footprint could be minimised by only loading part of the model in memory and always offloading the rest.
%
%
Moreover, while \tool{} supports splitting at any given layer, we limit the candidate split points of each network to the outputs of \texttt{ReLU} layers, due to their high compressibility (Section~\ref{sec:partitioning}). 
Although offloading could happen at sub-layer, filter-level granularity, this would impose extra overhead on the scheduler due to the significantly \mbox{larger search space}.
%


Our workflow also assumes the model to be available at both the client and server side.
While cloud resources are often dedicated to specific applications, edge resources tend to present locality challenges. To handle these, we could extend \tool{} to provide incremental offloading~\cite{Jeong2018} and cache popular functionality~\cite{Wang:2016:CCC:2984356.2988518} closer to its users.
%
%
In the future, we intend to explore multi-client settings and simultaneous asynchronous inferences on a single memory copy of the model, as well as targeting regression tasks and recurrent neural networks.


%% file: sections/conclusion.tex

\section{Conclusion}
\label{sec:conclusion}

In this paper, we present \tool{}, a distributed progressive inference engine that addresses the challenge of partitioning CNN inference across device-server setups. Through a run-time scheduler that jointly tunes the early-exit policy and the partitioning scheme, the proposed system supports complex performance goals in highly dynamic environments while simultaneously guaranteeing the robust operation of the end system. By employing an efficient multi-objective optimisation approach and a CNN-specific communication optimiser, \tool{} is able to deliver higher performance over the state-of-the-art systems across diverse settings, without sacrificing the overall system's accuracy and availability. 
 